\documentclass[conference]{IEEEtran}
\IEEEoverridecommandlockouts
\usepackage{times}

\usepackage[numbers,sort]{natbib}
\usepackage{multicol}
\usepackage[bookmarks=true]{hyperref}
\usepackage{graphicx} 
\usepackage{subfigure}
\usepackage{boldline}
\usepackage{amsmath}
\usepackage{mathtools}
\usepackage{multirow}
\usepackage{amsfonts}

\newcommand{\point}{\mathbf{p}}

\begin{document}

\title{iPlanner: Imperative Path Planning
\thanks{$^1$Fan Yang, Cesar Cadena, and Marco Hutter are with Robotic Systems Lab, ETH Zurich, 8092 Zürich, Switzerland. Emails: {\tt \{fanyang1, cesarc, mahutter\}@ethz.ch}}
\thanks{$^2$Chen Wang is with Spatial AI and Robotics Lab, State University of New York at Buffalo, NY 14260, USA. Email: {\tt chenwang@dr.com}}
}
\author{Fan Yang$^{1}$, Chen Wang$^{2}$, Cesar Cadena$^{1}$, and Marco Hutter$^{1}$}




%

\maketitle

\begin{abstract}
The problem of path planning has been studied for years. Classic planning pipelines, including perception, mapping, and path searching, can result in latency and compounding errors between modules. While recent studies have demonstrated the effectiveness of end-to-end learning methods in achieving high planning efficiency, these methods often struggle to match the generalization abilities of classic approaches in handling different environments. Moreover, end-to-end training of policies often requires a large number of labeled data or training iterations to reach convergence. In this paper, we present a novel Imperative Learning (IL) approach. This approach leverages a differentiable cost map to provide implicit supervision during policy training, eliminating the need for demonstrations or labeled trajectories. Furthermore, the policy training adopts a Bi-Level Optimization (BLO) process, which combines network update and metric-based trajectory optimization, to generate a smooth and collision-free path toward the goal based on a single depth measurement. The proposed method allows task-level costs of predicted trajectories to be backpropagated through all components to update the network through direct gradient descent. In our experiments, the method demonstrates around 4$\times$ faster planning than the classic approach and robustness against localization noise. Additionally, the IL approach enables the planner to generalize to various unseen environments, resulting in an overall 26-87$\%$ improvement in SPL performance compared to baseline learning methods.
\end{abstract}

\IEEEpeerreviewmaketitle


\section{Introduction}

Path planning is one of the significant tasks in the field of robotics. In structured environments, deploying a pre-built high-quality (HQ) map has pushed the limit and enabled robots to conduct daily tasks in environments shared with humans, e.g., autonomous driving~\cite{bao2022high}. However, for environments without a pre-built map, the robots can only rely on their onboard sensors for navigation. With limited sensor range, scene occlusions, and dynamic changes, the robots must react fast and re-plan efficiently to avoid collisions and navigate safely to their destinations.

The classic planning pipeline is based on a modular framework that includes perception, mapping, and path searching~\cite{wellhausen2021rough, cao2022autonomous}. This setup introduces latency as information is shared between modules and results in slower system response times. The planning performance is also limited by the information processed and preserved by each module, such as post-filtering and low-resolution perception. Maintaining a high-resolution map, on the other hand, can be computationally intensive and affect real-time performance. Additionally, if one module has errors or inaccuracies, these issues can be amplified and cause compounding effects by subsequent modules, potentially leading to system failure.

Researchers have been exploring end-to-end learning methods~\cite{shah2022gnm, wijmans2019dd} that can directly map sensory observations into trajectories or actions, as a way to improve the planning pipeline. \citet{shah2022gnm} trains the planning policies to mimic reference trajectories using supervised learning. However, this method relies on having diverse data to function well in various scenarios, which is a common limitation of explicit supervised training. On the other hand, non-supervised approaches like reinforcement learning (RL) \cite{wijmans2019dd}, which allows for random exploration, can lead to better generalizability. But, it struggles with low sampling efficiency and slow convergence, particularly in planning tasks with sparse task-level rewards. To address these issues, \citet{pfeiffer2018reinforced} combined supervised learning and RL, using prior demonstrations to improve training efficiency through better weight initialization using prior demonstrations. Recent studies~\cite{zhu2020vision, ye2021auxiliary} have utilized auxiliary tasks to boost RL's training efficiency with external supervised signals. However, this approach still faces the risk of overfitting to the training data due to the use of explicit training labels. Additionally, the auxiliary losses, which may not align with the main navigation objective, can steer the policy toward suboptimal solutions.

\begin{figure}[t]
    \vspace{0.1in}
    \centering
    \includegraphics[width=0.95\linewidth]{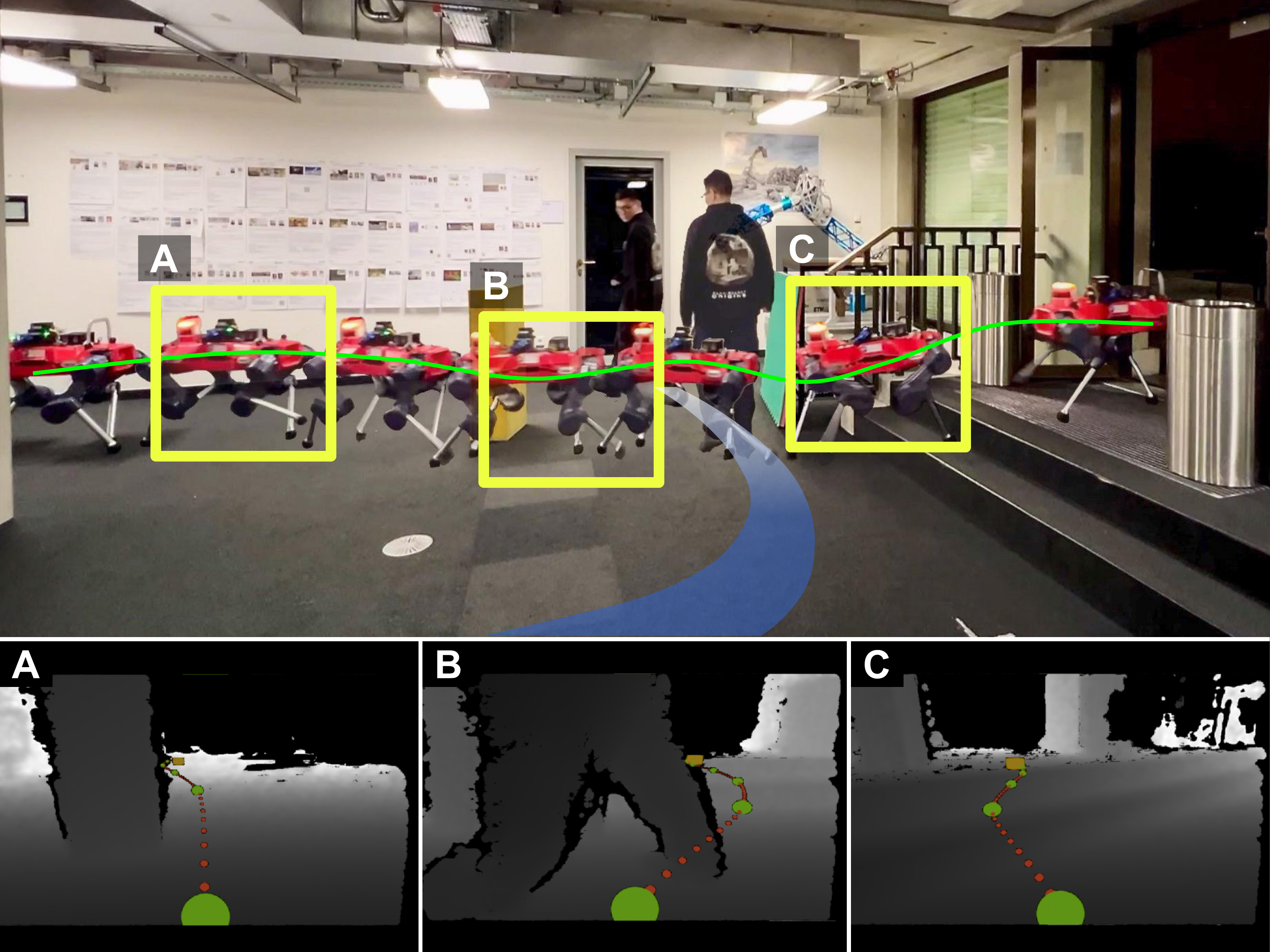}
    \caption{Experiment of planning through static and dynamic obstacles with a legged robot, with A, B, and C representing three planning events. The goal is set outside the door from the start. The green curve shows the robot's trajectory as it (A) avoids a static obstacle, (B) avoids a moving human, and (C) climbs stairs to reach the goal. The blue curve represents human movement. The bottom images illustrate the depth measurements and the predicted trajectories from our method for these three events.}
    \label{fig:opening_fig}
\end{figure}

\begin{figure*}
    \centering
    \includegraphics[width=1.0\linewidth]{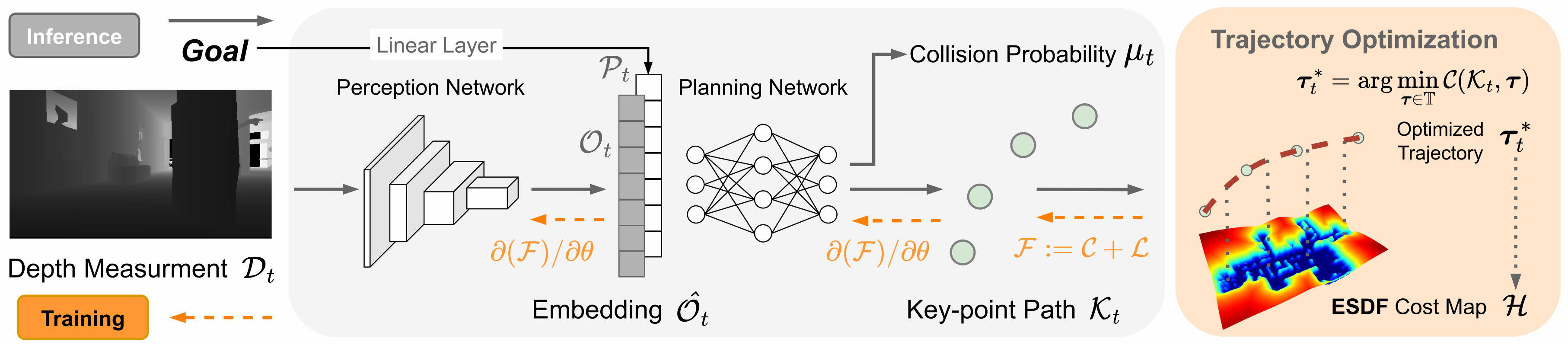}
    \caption{An overview of training the planning policy using IL. The pipeline consists of two parts, forming a BLO process with upper-level network update and lower-level trajectory optimization. During inference, the perception and planning network first encodes the depth measurement and goal position to predict a key-point path toward the goal with an associated collision probability. During training, the trajectory cost and task-level "fear" loss are propagated back to provide direct gradients for updating both the perception and planning networks simultaneously.}
    \label{fig:viplanner_diagram}
\end{figure*}

To address this challenge, we introduce ``imperative learning" (IL), a non-supervised approach aimed at improving the training efficiency and generalization of the policy. The policy includes network prediction and metric-based optimization, forming a BLO process during training. The IL approach trains the policy end-to-end without needing demonstrations or labeled references. Fig.~\ref{fig:viplanner_diagram} illustrates the overall process of the IL-based training for the planning policy. The core of IL lies in using differentiable metric-based optimization to direct network updates, resulting in an unsupervised ``imperative" loss function. During training, a pre-built differentiable cost map provides traversability cost to guide the policy's behavior. When deployed, the policy decodes the traversability information directly from the input ego-centric observation and plans on top of it. Due to end-to-end training, the observation features extracted during deployment can be optimized specifically for the planning objective.
Overall, the proposed IL approach can offer advantages over both end-to-end RL and supervised learning methods. In comparison to the end-to-end RL, instead of stochastically sampling the policy, the pre-computed cost map can guide the optimization process through direct gradient descent, improving training efficiency. In contrast to supervised learning, the IL approach utilizes task-level loss without any explicit labels or demonstrations, resulting in increased exploration of the action space, leading to better generalization. This paper presents the IL approach as a novel solution to overcome the limitations of previous methods for training a perceptive planning policy. Its main contributions are summarized as follows:
\begin{itemize}
  \item A new non-supervised learning approach to train a perceptive planning policy with ``imperative" supervision through direct gradient descent.
  \item An end-to-end training pipeline to map the single depth measurement to the trajectory by leveraging a BLO process with network update and metric-based trajectory optimization.
  \item Benchmarking the planning performance of learning-based methods with the classic approach in handling different types of unseen environments.
\end{itemize}

The iPlanner will be open-sourced\footnote{iPlanner: \url{https://github.com/leggedrobotics/iPlanner}} to promote research for learning navigation autonomy.


\section{Related Work}
Nowadays, classic planning frameworks, such as those done by \citet{yang2021real, wellhausen2021rough, cao2022autonomous}, have demonstrated effective and reliable performance in autonomous navigation in unseen terrains. For example, the navigation system, done by \citet{cao2022autonomous}, incorporating mapping, terrain analysis, and motion-primitive path search\cite{zhang2020falco}, is deployed successfully during the DARPA Subterranean Challenge. The work of \citet{wellhausen2021rough} employs a pipeline with a learned terrain estimator and executes a lazy-PRM~\cite{kavraki2000path} path search, similar to the approach of \citet{yang2021real} who uses a learned terrain cost and performs PRM*~\cite{karaman2011sampling} with additional path optimization. Recently, there has been a shift towards end-to-end policies that directly map sensory inputs to planning paths or control actions~\cite{pfeiffer2017perception, bojarski2016end, loquercio2021learning, xiao2022learning, shah2022gnm, shi2019end, wijmans2019dd, zhu2020vision, ye2021auxiliary, kahn2021badgr, hoeller2021learning, kim2022learning, sadat2020perceive}. This is due to the challenges posed by modularized systems, such as increased overall latency~\cite{loquercio2021learning} and lack of robustness against noise or errors in real-world missions. However, the generalizability and reliability of these learning methods may still lag behind classic approaches, especially in scenarios that have not been encountered during training.

\textit{Supervised Learning}: A common method for training an end-to-end policy is to copy expert or ``ground truth" trajectories, such as works done by \citet{pfeiffer2017perception, bojarski2016end, loquercio2021learning, xiao2022learning, shah2022gnm, sadat2020perceive}. The policy is trained to associate input sensory data with ground truth labels, which are obtained either by demonstrations from experts or statistical sampling. However, the explicitly supervised policy may not generalize well to diverse environments because of limited data diversity and richness. \citet{shah2022gnm} uses 60 hours of navigation data from multiple robots and environments to improve generalization, but its performance is still limited by scene coverage. Moreover, relying on expert trajectories~\cite{sadat2020perceive} or explicitly labeled references may limit the performance of the planner, leading it to converge towards sub-optimal solutions due to the limitations of the optimalities of the expert paths. \citet{loquercio2021learning} used simulated environments to train a planning policy for fast-flying drones. The simulated depth is estimated using stereo matching to reduce the gap between simulation and reality, allowing the network to be exposed to more training data with less effort in real-world data collection. Nevertheless, the use of a sampling-based planning algorithm to generate reference paths may limit the performance of the network by the optimality of the sampling-planning results.

\textit{Reinforcement Learning}: Recently, deep reinforcement learning has gained popularity for training planning policies, as demonstrated in studies by \citet{shi2019end, wijmans2019dd, zhu2020vision, ye2021auxiliary, kahn2021badgr, hoeller2021learning, kim2022learning}. As mentioned above, RL can utilize task-level loss. With simulated environments, using RL can save the effort of collecting and labeling data and generate various scenarios to improve the generalization of the policy. Moreover, the unsupervised nature of RL does not depend on the optimalities of expert labels and allows the network to explore the action space for better solutions. However, there are challenges in training with RL. One issue is the difficulty in creating a perfect simulator. To overcome the sim-to-real gap, researchers may need to collect data in the real world~\cite{kahn2021badgr} or pre-process the simulated inputs based on heuristics~\cite{hoeller2021learning}. Additionally, RL's low sample efficiency makes it challenging to train large policies end-to-end from dense inputs such as images or depth measurements. Training may take several days even with multiple GPUs~\cite{wijmans2019dd}. To reduce the complexity of training, researchers have adopted strategies such as separating perception training from policy training~\cite{hoeller2021learning} or using auxiliary tasks with heuristics-based loss functions~\cite{zhu2020vision, ye2021auxiliary}. However, the success of these methods is contingent upon the accuracy and optimality of the heuristics employed.

This paper explores a novel non-supervised IL approach to train a planning policy from depth measurements end-to-end. This IL method uses task-level loss, eliminating the need for explicit labeling during training. By utilizing a differentiable cost map, the IL loss function can guide the network update through direct gradient descent to improve the training efficiency.



\section{Methodology}

\textit{Problem Definition}: Define $\mathcal{Q} \subset \mathbb{R}^3$ as the workspace for the robot to navigate. Let the subset $\mathcal{Q}_{\rm obs} \subset \mathcal{Q}$ represent the obstacles in the space that the robot cannot traverse through. Given a depth observation $\mathcal{D}_{t} \in \mathbb{R}^{H \times W}$ from the current time stamp $t$ and a goal $\point^{G}_t \in \mathbb{R}^3$ in the robot frame, a trajectory $\boldsymbol{\tau}_{t}$ can be generated to guide the robot from the current position $\point^{R}_t \in \mathbb{R}^3$ towards to the goal $\point^{\rm G}_t$ while avoiding obstacles $\mathcal{Q}_{\rm obs}$. Here, we also denote a cost function $\mathcal{C}$ of a given trajectory $\boldsymbol{\tau}$, and an overall task-level loss function $\mathcal{F}$, such that $\mathcal{F} = \mathcal{C} + \mathcal{L}$, where $\mathcal{L}$ is an arbitrary non-negative function.

\textit{System Overview}: The process of planning and imperative learning is illustrated in Figure~\ref{fig:viplanner_diagram}. The pipeline consists of three modules. Firstly, a convolutional neural network (CNN) based~\cite{lecun1995convolutional} perception front-end transforms the depth input into an observation embedding. This embedding, combined with the goal waypoint feature, is then input into a planning network that predicts a key-point path, $\mathcal{K}_t$, towards the goal, which consists of $n$ key points. Afterward, a metric-based trajectory optimizer further optimizes and interpolates the key-point path based on the cost $\mathcal{C}$ on the cost map $\mathcal{H}$ to generate the trajectory $\boldsymbol{\tau}_t$. The optimized cost $\mathcal{C}$, with other task loss $\mathcal{L}$, is then backpropagated through the network to update its parameters $\theta$. Together, the network update and trajectory optimization form a nested BLO process. The aim of the pipeline is to enforce a trajectory optimization-based supervision (unsupervised optimization) on the perception and planning network $f$ parameterized by $\theta$. We hereby formulate the BLO model as follows:
\begin{equation}\label{eq:objective}
    \begin{aligned}
    \displaystyle \min_{\theta} \: & \mathcal{F}(f_\theta, \boldsymbol{\tau}^*) \\
    \text{s.t.} \: & \boldsymbol{\tau}^* = \displaystyle \arg \min_{\tau\in \mathbb{T}} C(f_\theta, \boldsymbol{\tau}),
    \end{aligned}
\end{equation}
where $\boldsymbol{{\tau}}^*$ is the optimized trajectory under a constraint set $\mathbb{T}$, based on the objective $\mathcal{C}$.  We next discuss the intuition of \eqref{eq:objective} in detail.

\subsection{Perception and Planning Network}
\textit{Perception Network}: At each time stamp $t$, the robot receives a depth measurement $\mathcal{D}_{t}$ sampled from the space $\mathcal{Q}$. The front-head perception network encodes the observation $\mathcal{D}_{t}$ from the time $t$ into a perception embedding $\mathcal{O}_{t} \in \mathbb{R}^{C \times M}$, preserving spatial and geometric information for the planning. Our method utilizes the backbone of the ResNet, proposed by \citet{he2016deep}, a classic and popular CNN architecture that is widely used as a feature extractor for image data. We adopt the efficient architecture: ResNet-18, to extract a feature representation for path planning from the perception input $\mathcal{D}_{t}$.

\textit{Planning Network}: The planning network takes the perception embedding $\mathcal{O}_{t} \in \mathbb{R}^{C \times M}$, derived from the depth measurement $\mathcal{D}_{t}$, and the goal position $\point^{\rm G}_t \in \mathbb{R}^3$ to find an collision-free to the goal. Here, $C$ denotes the channel dimension of the embedding $\mathcal{O}_{t}$, and $M$ represents the dimension of the feature space. The goal position, with a dimension of $3$, is first mapped into a higher dimensional feature embedding $\mathcal{P}_{t} \in \mathbb{R}^{C^* \times M}$, where $C^* \geq 3$ using a linear layer. The perception embedding $\mathcal{O}_{t}$ and the expanded goal feature $\mathcal{P}_{t}$ are concatenated to form $\mathcal{\hat{O}}_{t} \in \mathbb{R}^{(C+C^*)\times M}$ as the observation input. The planning network uses a combination of CNN and MLP with ReLU activation to process $\mathcal{\hat{O}}_{t}$ and predict a key-point path $\mathcal{K}_t \in \mathbb{R}^{n \times 3}$ containing $n$ key points. The path $\mathcal{K}_t$ is then interpolated and optimized by a metric-based trajectory optimizer.


\subsection{Trajectory Optimization (TO)}
To enforce the safety and smoothness of the trajectory generated by the planner, the key-point path $\mathcal{K}$ predicted by the planning network, is optimized by a trajectory optimizer. The objective is to minimize the trajectory cost $\mathcal{C}$ on a pre-defined cost map $\mathcal{H}$, which will be discussed in more detail later. The optimization takes the predicted key-point path $\mathcal{K}$ as the initial input and output the trajectory $\boldsymbol{\tau}^*$ under a constraint set $\mathbb{T}$, formulated as follow:
\begin{equation}
    \begin{aligned}
    \displaystyle \boldsymbol{\tau}^*  = \arg \min_{\boldsymbol{\tau} \in \mathbb{T}} \mathcal{C}(\mathcal{K}, \boldsymbol{\tau}), \,\, \text{where} \;  \mathcal{K} \leftarrow f_\theta
    \end{aligned}
\end{equation}
Notice that the pre-built cost map is utilized only during training. During deployment, the TO interpolates and smooths the key-point path to generate a dynamically feasible trajectory $\boldsymbol{\tau}_{t}$ under a system constraint. Specifically, here, we use Cubic-Spline~\cite{mckinley1998cubic} to construct a third-order polynomial to pass through key points as a constraint. It interpolates $m$ intermediate points in between every two key points in $\mathcal{K}_t$ and generates a dynamically feasible trajectory $\boldsymbol{\tau}_{t} \in \mathbb{R}^{(mn+1)\times3}$ that passes through all the key points, with zero second derivatives on the start and endpoints, as per the natural boundary conditions. This optimization problem can be solved as a symmetric tridiagonal system with a linear solution~\cite{bartels1995introduction}. Here, we create the cubic-spline function as a differentiable layer using PyPose~\cite{wang2023pypose} library to record the gradients of the output trajectory $\boldsymbol{\tau}_{t}$ to the input constraints $\mathcal{K}_t$ predicted by the network.

\begin{figure}
    \centering
    \subfigure[Environment Point Cloud \label{fig:env_pointcloud}]{\includegraphics[height=0.3\linewidth]{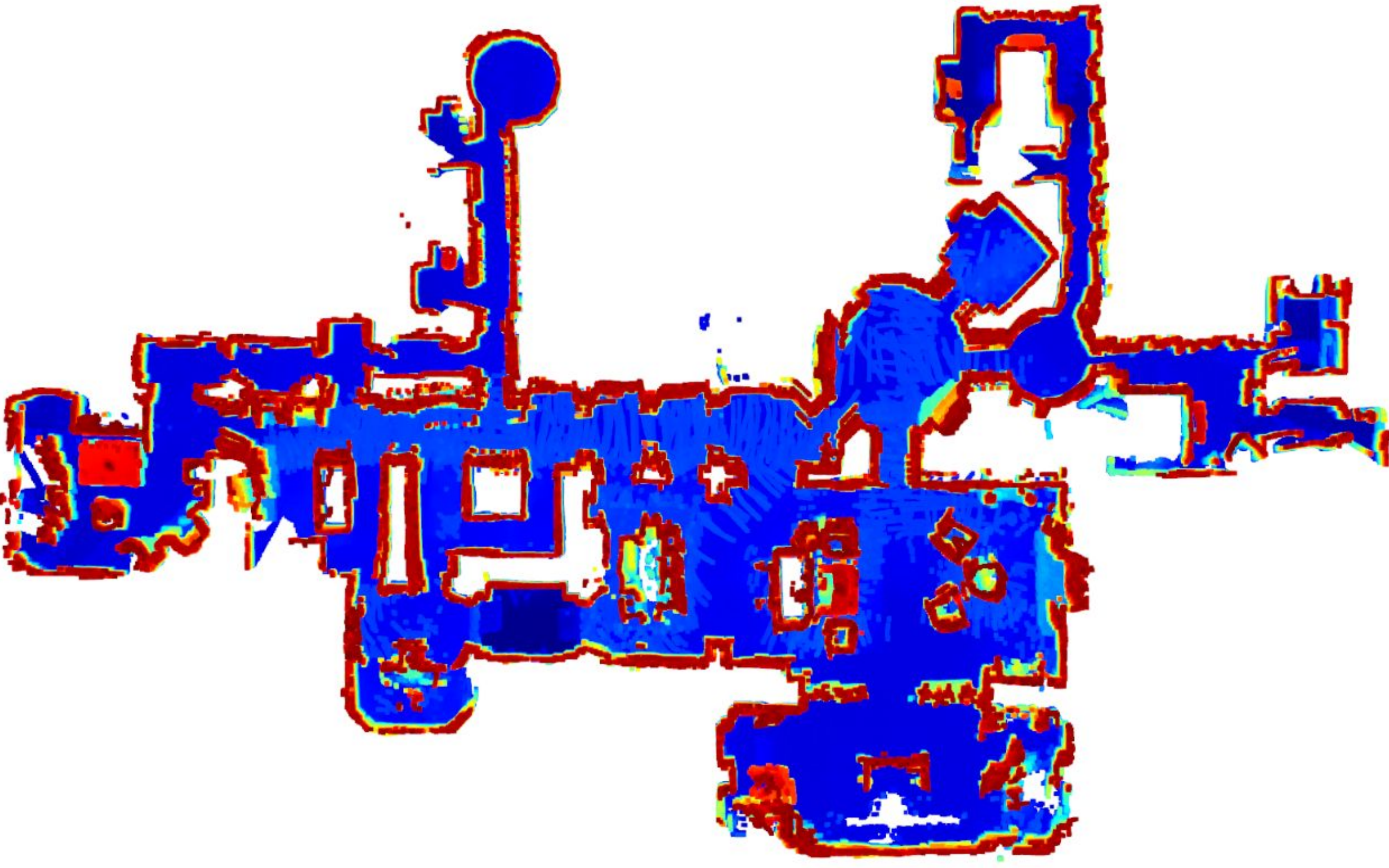}} 
    \subfigure[ESDF Cost Map \label{fig:esdf}]{\includegraphics[height=0.3\linewidth]{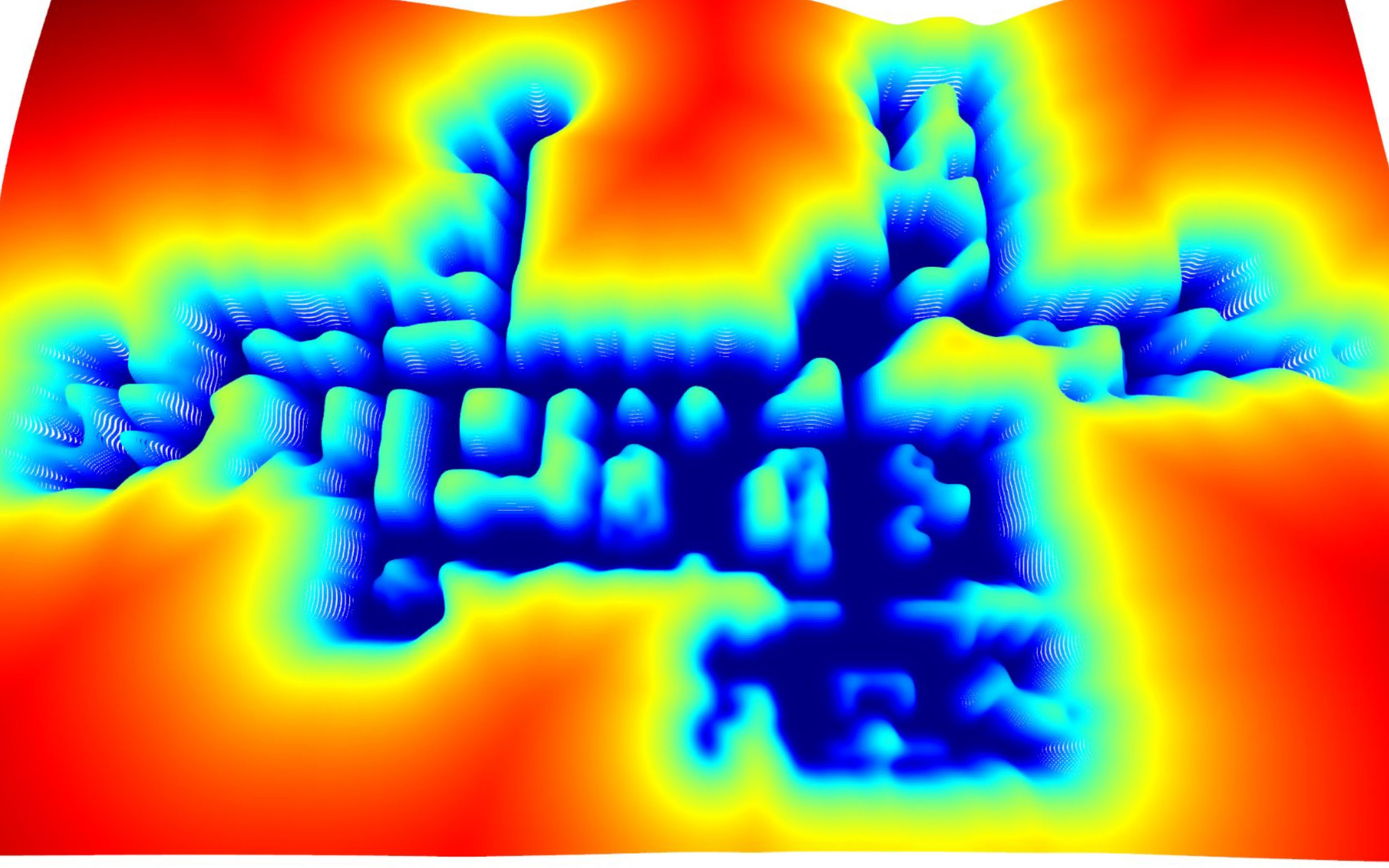}}
    \caption{An illustration of a training environment and its ESDF cost map generated from the Matterport3D~\cite{Matterport3D} dataset. (a) depicts the point cloud reconstructed from collected depth images within a Matterport3D room. (b) shows the smoothed ESDF cost map produced from the point cloud with Gaussian filtering.}
    \label{fig:esdf_map}
\end{figure}

\begin{figure}
    \centering
    \includegraphics[width=0.95\linewidth]{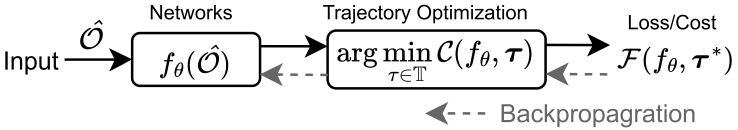}
    \caption{The mathematically BLO pipeline of the imperative training for the planning policy. The network function $f_\theta$ predicts a path as the input for the TO process. The TO minimizes the cost $\mathcal{C}$ with the optimized trajectory $\boldsymbol{\tau^*}$. The trajectory cost $\mathcal{C}$, combined with additional loss terms, forms the total training loss $\mathcal{F}$, and $\mathcal{F}$ is then backpropagated to update the network parameter $\theta$.}
    \label{fig:backward_process}
\end{figure}

\subsection{Optimization Objective and Training Loss}
\subsubsection{Trajectory Cost}The trajectory cost includes three differentiable terms that assess the quality of the output trajectory $\boldsymbol{\tau}_{t}$. The first term, obstacle cost $\mathcal{C^O}$, checks if the trajectory collides or is too close to obstacles. To this end, before training, we reconstruct the environment offline based on the collected depth images, and their associated camera poses, as shown in Fig.~\ref{fig:env_pointcloud}. Then, based on the reconstructed environment, we build a Euclidean Signed Distance Field (ESDF) to label the distance to the nearest free (non-obstacle) space for each position in the environment. The ESDF is then smoothed with a Gaussian filter to make it locally differentiable and create a cost map $\mathcal{H}$ with non-negative cost values, shown in Fig.~\ref{fig:esdf}. To generate the cost $\mathcal{C^O}$, each coordinate $\point_i$ on the trajectory $\boldsymbol{\tau}_{t}$ will be projected on this cost map $\mathcal{H}$ to get a cost value. The formulation of the obstacle cost $\mathcal{C}^{O}$ is shown as follows:
\begin{equation}
    \begin{aligned}
    \mathcal{C^O}(\boldsymbol{\tau}_{t}) = & \sum_{i}\mathcal{H}(\point_i) & \point_i \in \boldsymbol{\tau}_{t}, |_{i = m \times n},\\
    \end{aligned}
\end{equation}
where $\mathcal{H}(\point_i)$ represents the value of position $\point_i$ on the cost map $\mathcal{H}$. Secondly, we use the Euclidean distance from the final position of the trajectory $\boldsymbol{\tau}_{t}$ to the goal $\point^G_t$ as the second cost term $\mathcal{C^G}$. It encourages the trajectory to be close to the destination and punishes the one that deviated away. The destination cost $\mathcal{C^G}$ is hereby denoted as follow:
\begin{equation}
    \begin{aligned}
    \mathcal{C^G}(\boldsymbol{\tau}_{t}) & = \mathnormal{E}(\point_i, \point^G_t) & \point_i \in \boldsymbol{\tau}_{t} \, |_{i = m \times n},\\
    \end{aligned}
\end{equation}
where function $\mathnormal{E}$ calculates the Euclidean distance between two given positions. Finally, the cost term $\mathcal{C^M}$ is used to evaluate the motion smoothness of the trajectory. We assume that the same amount of time is taken to travel between two key points $\point^K_i, \point^K_{i+1} |_{i=0...n-1}$ on the key-point path $\mathcal{K}_t$ and perform equal-time interpolation during the TO process. The objective of minimizing the cost $\mathcal{C^M}$ is to reduce the difference in length of trajectory intervals on $\boldsymbol{\tau}_t$, thereby minimizing overall acceleration. The Euclidean distance from the current robot position $\point^R_t$ to the goal $\point^G_t$ is used to regulate the overall trajectory length and rewards a shorter motion. The motion cost $\mathcal{C^M}$ is formulated as follow:
\begin{equation}
    \begin{aligned}
    \mathcal{C^M}(\boldsymbol{\tau}_{t}, \point^R_t, \point^G_t) & = \sum_{i=0}^{n-2}\mid\frac{\mathnormal{E}(\point^R_t, \point^G_t)}{n-1} \, - \mathnormal{E}_{\boldsymbol{\tau}_t}(\point^K_i, \point^K_{i+1})\mid,
    \end{aligned}
\end{equation}
where function $\mathnormal{E}_{\boldsymbol{\tau}_t}$ returns the length of a path interval between two given positions on the trajectory $\boldsymbol{\tau}_t$. The trajectory cost $\mathcal{C}$, as the objective of TO, is then formulated as a combination of $\mathcal{C^O} \in \mathbb{R}^{+}_0$, $\mathcal{C^G} \in \mathbb{R}^{+}$, and $\mathcal{C^M} \in \mathbb{R}^{+}$:
\begin{equation}
    \begin{aligned}
    \mathcal{C}(\boldsymbol{\tau}_{t}) = & \alpha \, \mathcal{C^O}(\boldsymbol{\tau}_{t}) + \beta \, \mathcal{C^G}(\boldsymbol{\tau}_{t}) + \gamma \, \mathcal{C^M}(\boldsymbol{\tau}_{t}, \point^R_t, \point^G_t),
    \end{aligned}
\end{equation}
where $\alpha$, $\beta$, and $\gamma \in \mathbb{R}^{+}$ are hyperparameters used to balance the cost between different terms.

\subsubsection{Fear Loss} In addition to the trajectory cost, the planning network predicts a collision probability $\mu_t$ for each trajectory to assess its risk of collision with obstacles. This is referred to as ``fear loss". In some scenarios, the local planning policy can be trapped in a local minimum. Instead of setting large obstacle costs that could result in an over-conservative policy, having this task-level loss in addition to the trajectory cost provides the planner with the flexibility to escape from those scenarios and ensures safety. When deploying, the planner will only execute the trajectory with associated collision probability $\mu_t < 0.5$. The fear loss $\mathcal{L}(\boldsymbol{\tau}_t)$ is calculated using binary cross entropy (BCE):
\begin{equation}
    \mathcal{L}(\boldsymbol{\tau}_t) = \left\{
    \begin{aligned}
    &\text{BCELoss}(\,\mu_t, \,1.0\,)   & \boldsymbol{\tau}_t \; \text{collides w.}\, \mathcal{Q}_{\rm obs} \\
    &\text{BCELoss}(\,\mu_t, \,0.0\,)  & \text{otherwise}. \\
    \end{aligned}
    \right.
\end{equation}

The final training loss $\mathcal{F}$ is then formulated as the general summation of trajectory cost $\mathcal{C}$ and task-level fear loss $\mathcal{L}$:
\begin{equation}
    \begin{aligned}
    \mathcal{F} = \mathcal{C}(\boldsymbol{\tau}_t) + \mathcal{L}(\boldsymbol{\tau}_t).
    \end{aligned}
\end{equation}

\begin{figure}
    \centering
    \subfigure[]{\includegraphics[height=0.3\linewidth]{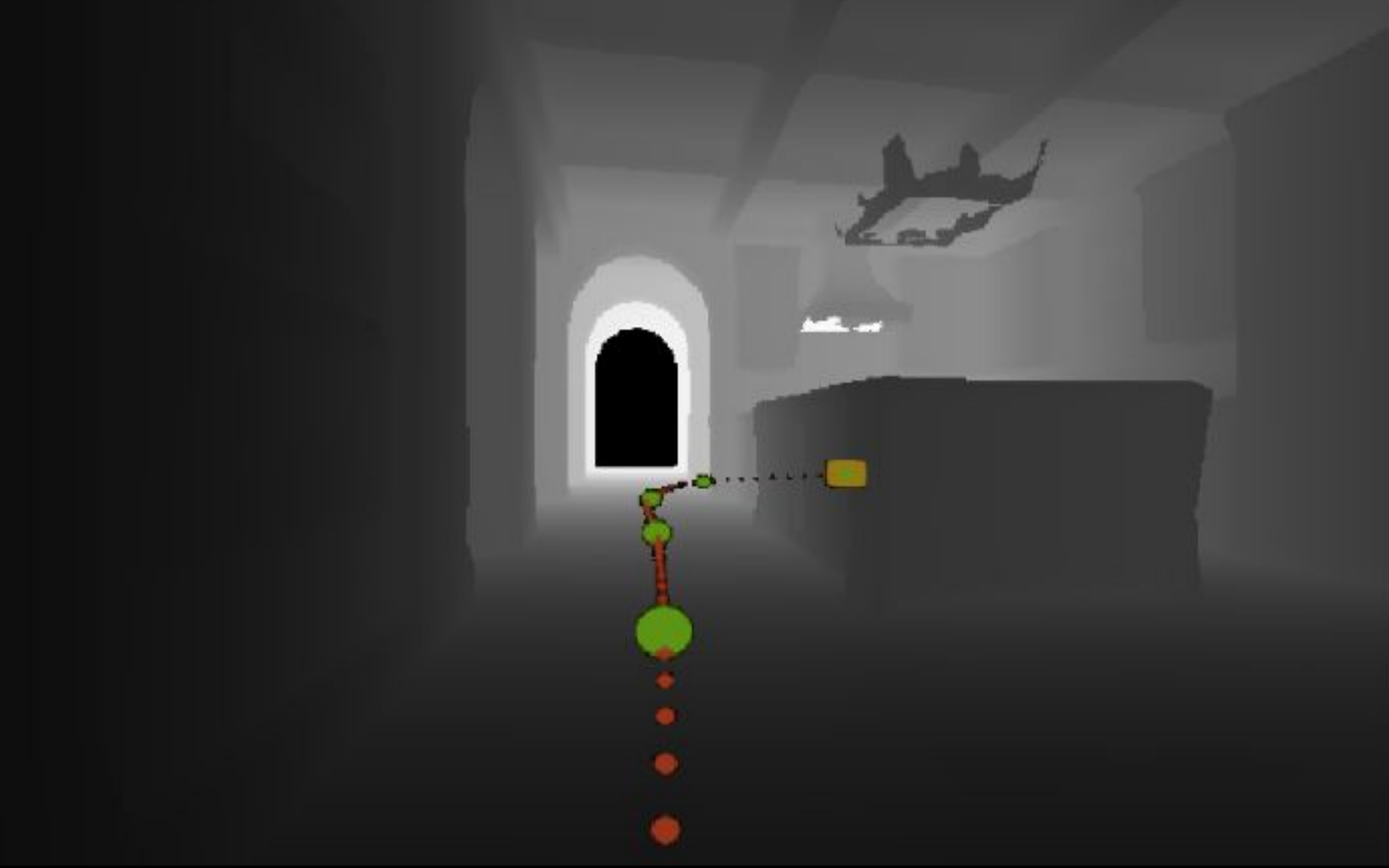}} 
    \subfigure[]{\includegraphics[height=0.3\linewidth]{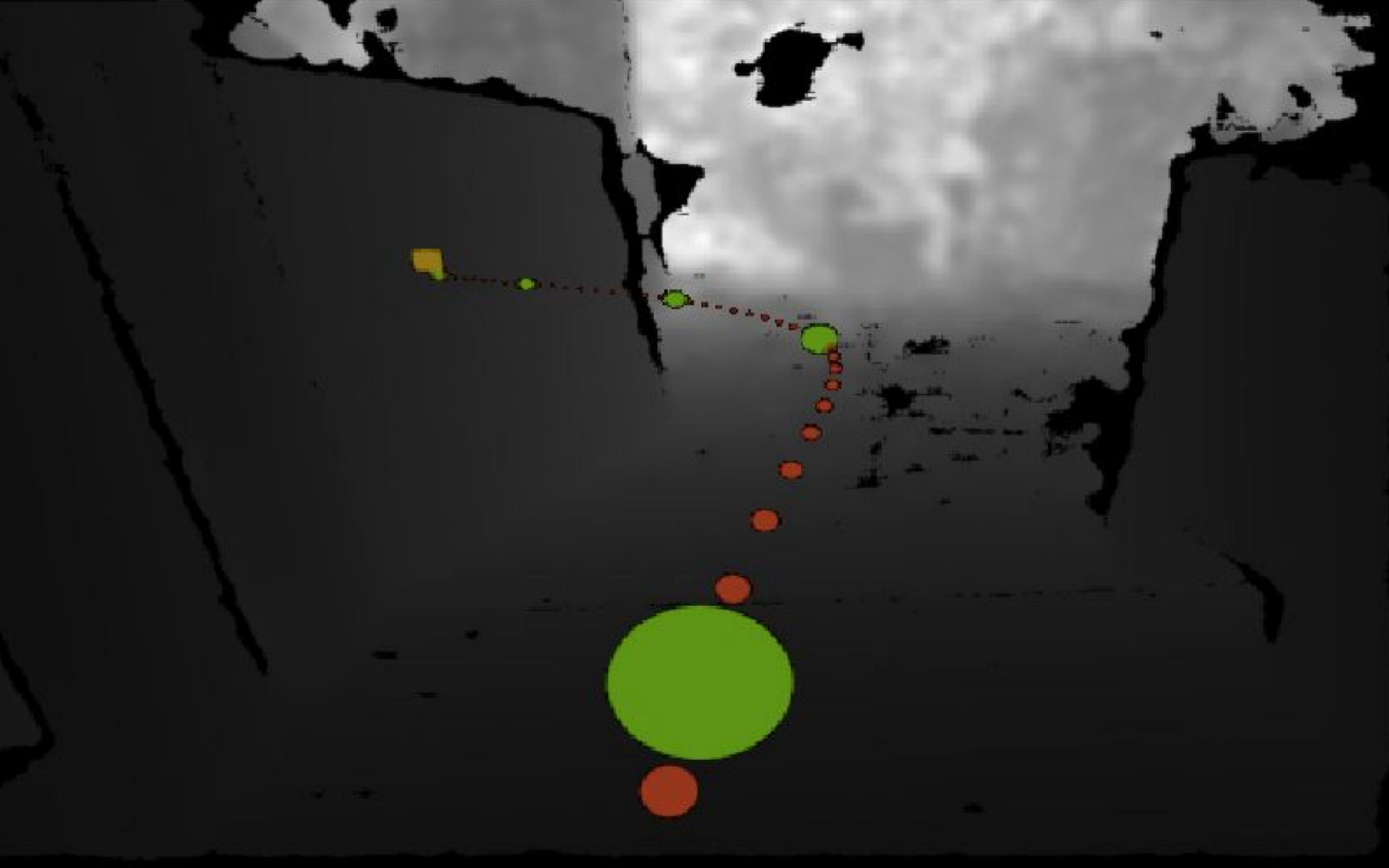}} \\
    \caption{Illustration of depth observation from simulation and the real world. (a) A depth image generated from the Gazebo simulation in the Matterport3D environment. (b) A depth observation obtained from Intel RealSense D435 during real-world experiments.}
    \label{fig:image_examples}
\end{figure}

\begin{figure}
    \centering
    \includegraphics[width=0.98\linewidth]{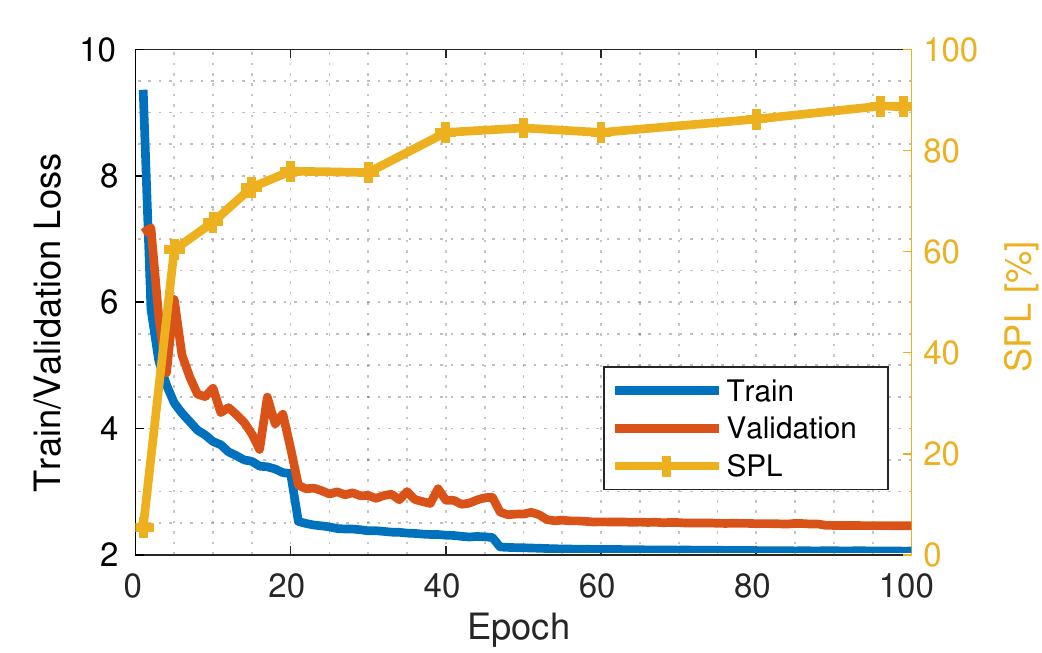}
    \caption{Example of training and validation loss throughout the training process. The SPL (Success Weighted by Path Length) is evaluated in the garage simulation environment after specific epochs of training.}
    \label{fig:training_plot}
\end{figure}

\begin{figure*}
    \centering
    \subfigure[Indoor]{\includegraphics[height=0.31\linewidth]{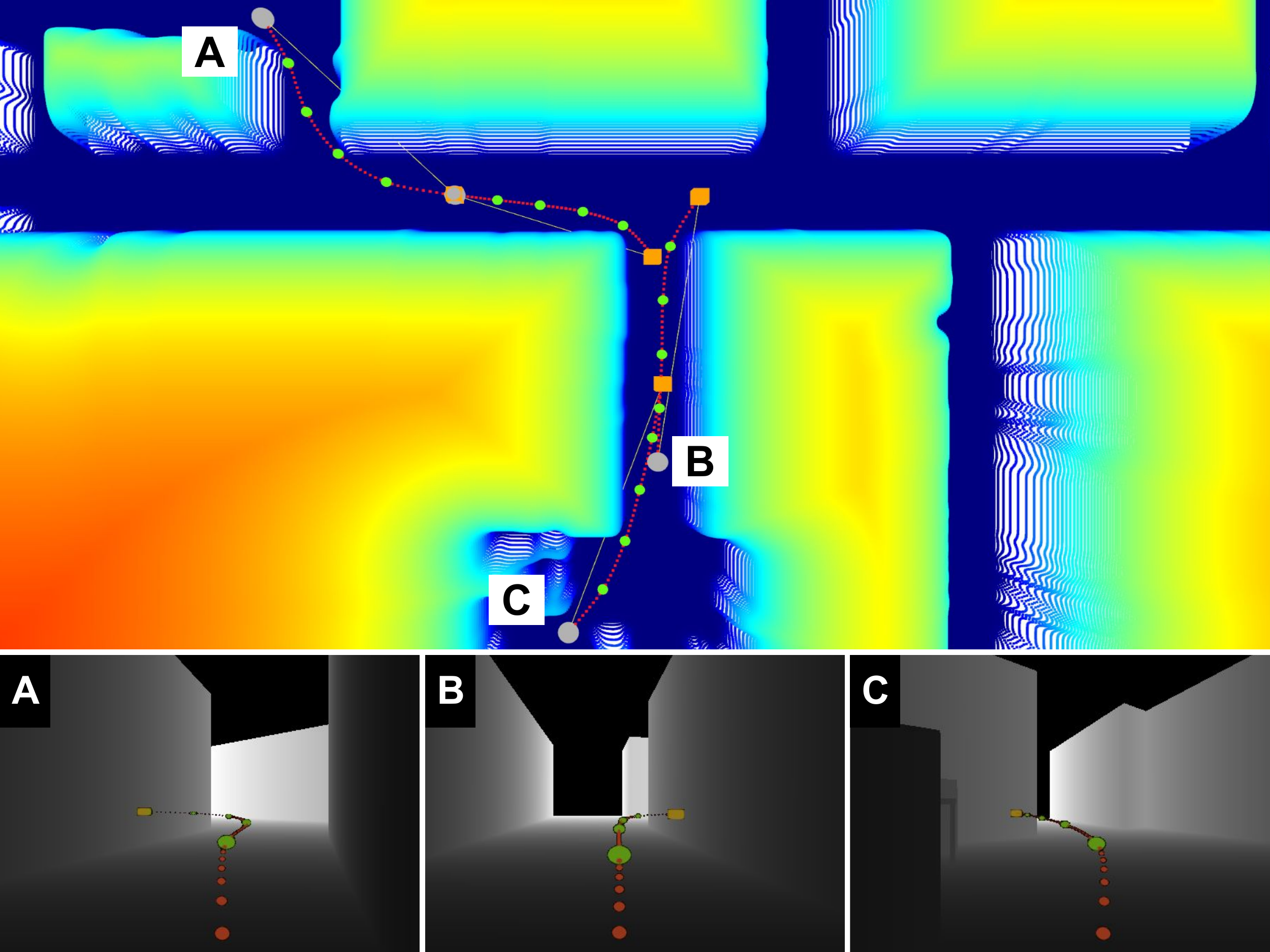}} \hspace{0.1in}
    \subfigure[Garage]{\includegraphics[height=0.31\linewidth]{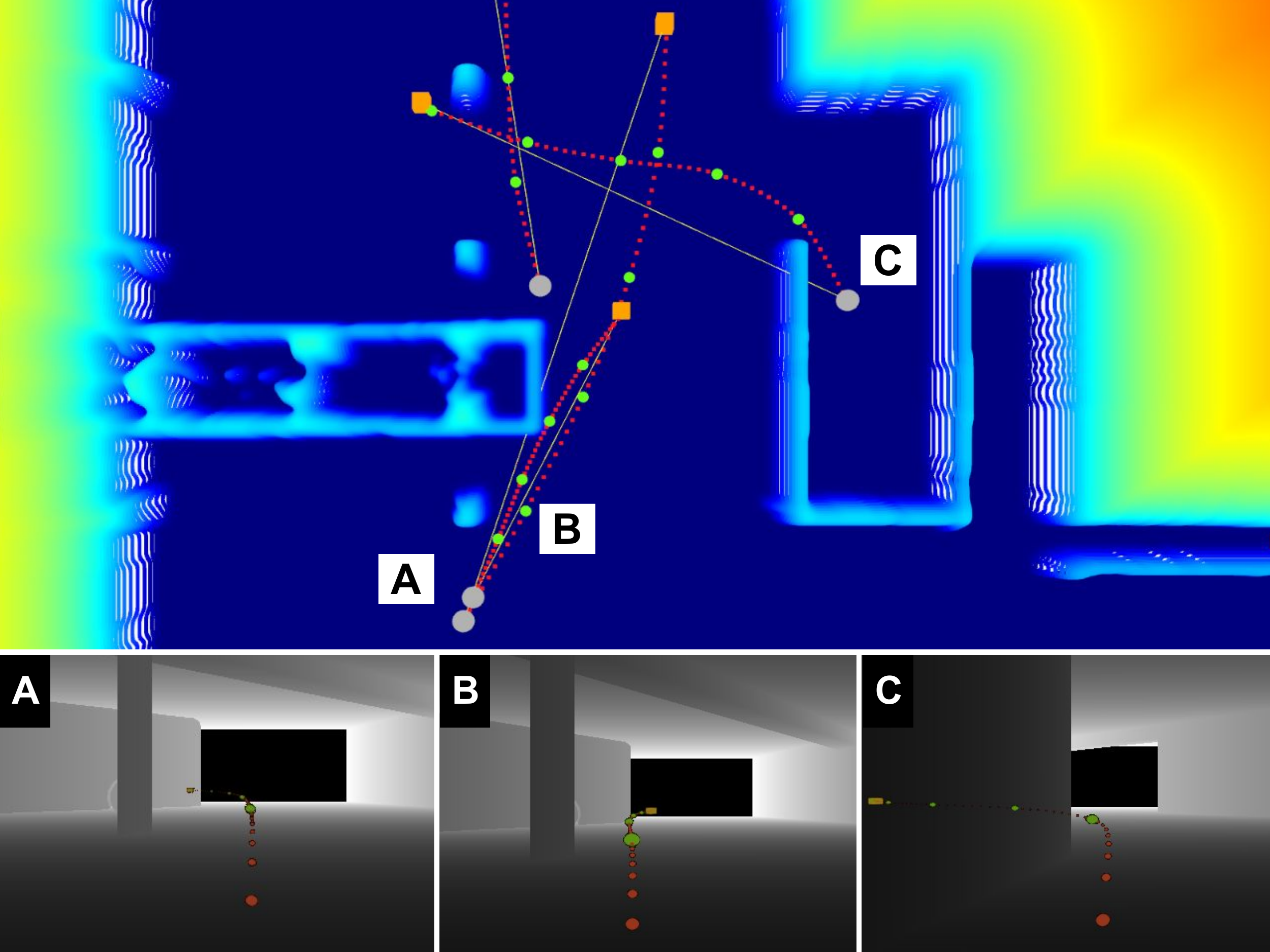}} 
    \subfigure[Forest]{\includegraphics[height=0.31\linewidth]{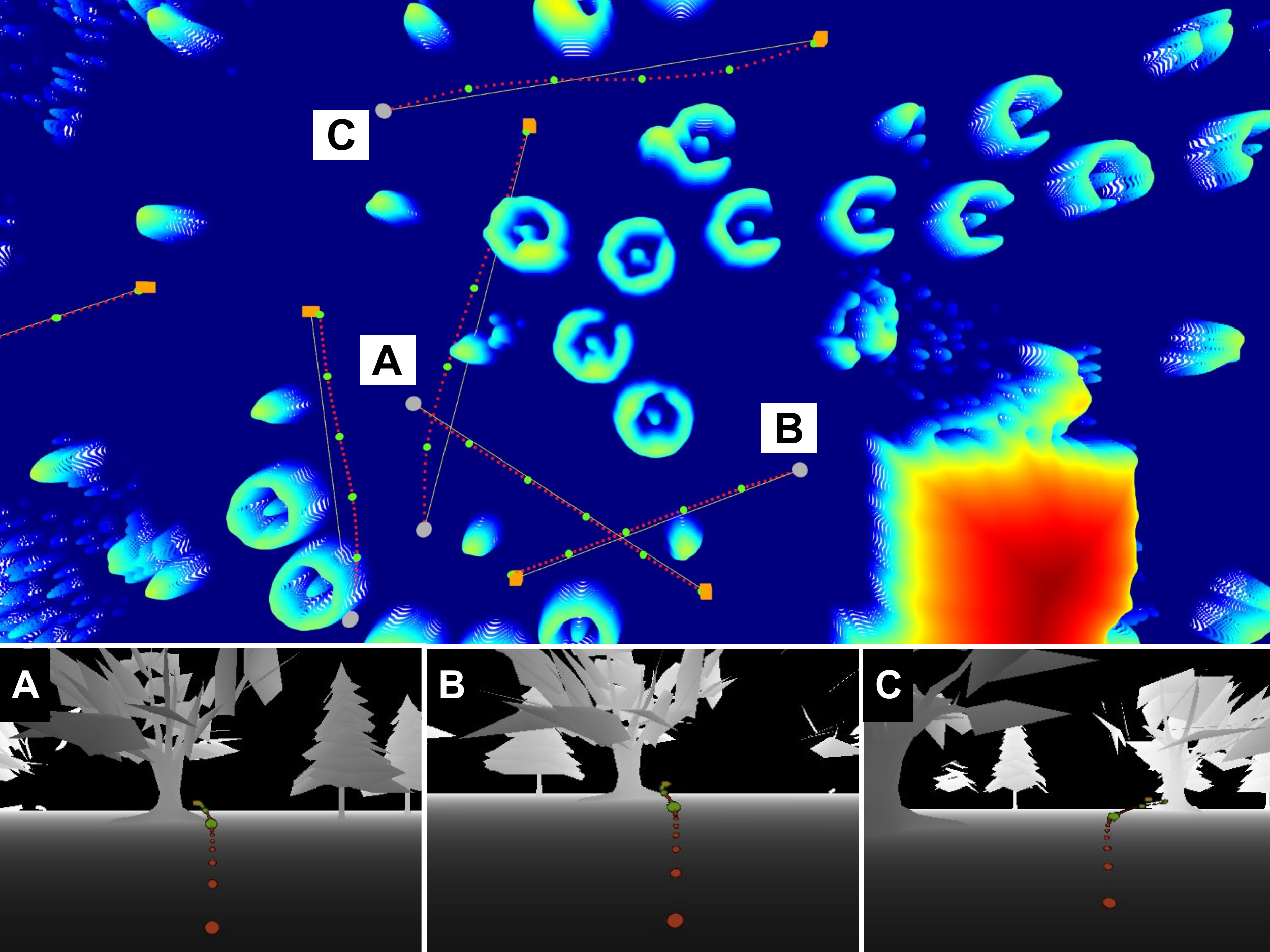}} \hspace{0.1in}
    \subfigure[Matterport]{\includegraphics[height=0.31\linewidth]{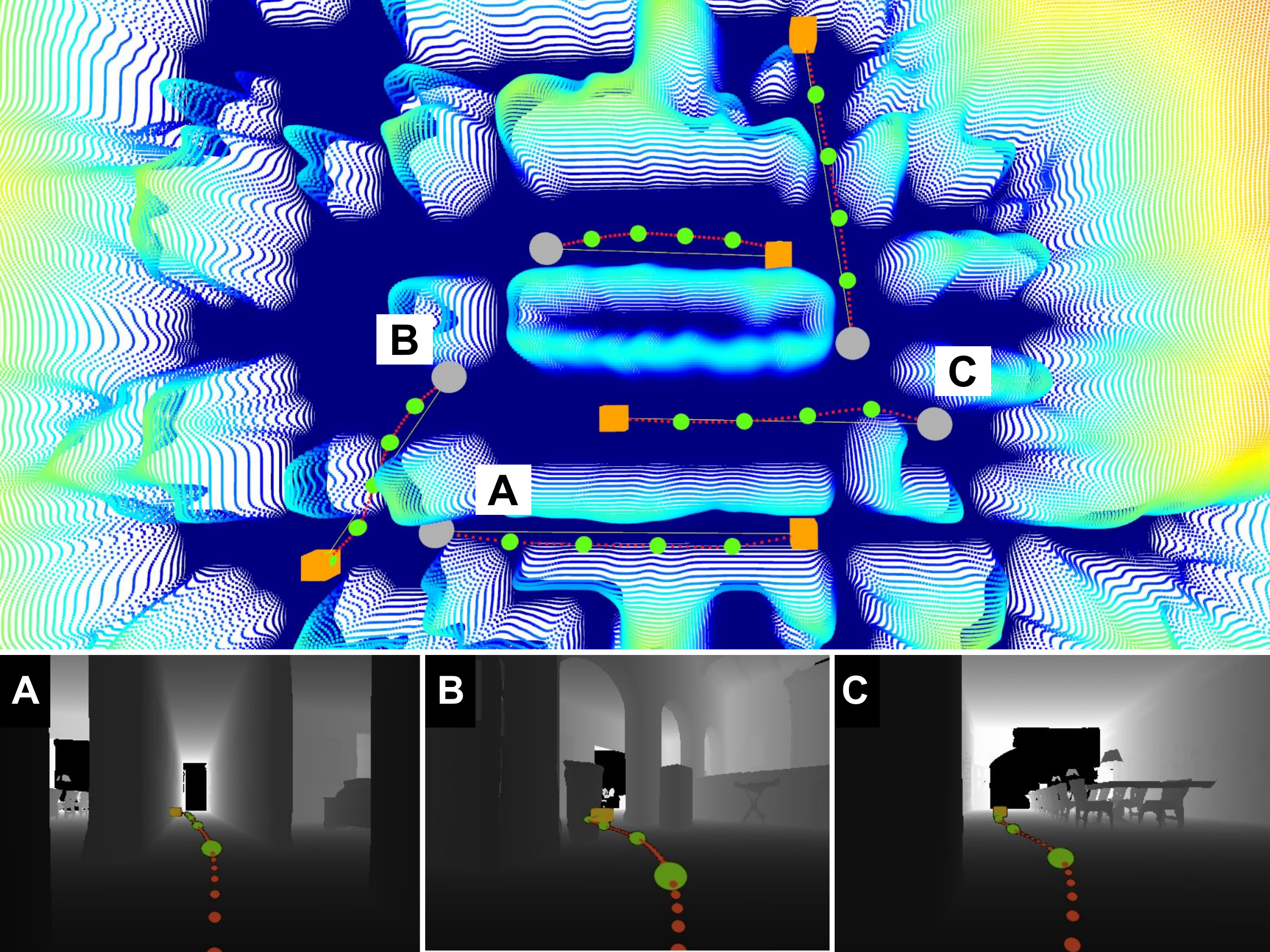}}
    \caption{Planning examples in simulation experiments: (a) Indoor, (b) Garage, (c) Forest, (d) Matterport3D. The bottom images show the depth measurements for the corresponding marked trajectories on the cost maps. The key points (green), along with the trajectories (red) shown in the depth images, or on cost maps, are generated by our method from the current robot positions (gray dot) to the goals (yellow/orange box), with the current depth images. The trajectories shown in the depth measurements are projected onto the image frame for visualization purposes.}
    \label{fig:planning_examples}
\end{figure*}

\subsection{Training and Bi-Level Optimization (BLO)} As described above, the pipeline of the IL forms a nested BLO process, where the TO, as the lower-level task, takes the output of the network to generate the trajectory $\boldsymbol{\tau}$. Then, given the output $\boldsymbol{\tau}$, the upper-level process of BLO finds the network parameter $\theta$ to minimize the final training loss $\mathcal{F}$. Fig.~\ref{fig:backward_process} shows the pipeline in a mathematical formulation. To solve the BLO problem, we adopt the concept of Proxy-Based Explicit Gradient for Best-Response (EGBR)~\cite{liu2021investigating} by using approximated BR mapping. Specifically, instead of explicitly solving the optimization of the lower-level task TO, which can be computationally expensive, we approximate a sub-optimal solution $\boldsymbol{\tau}^{\text{sub}}$ by directly using the output of the cubic-spline interpolation and its projection cost on map $\mathcal{H}$ as the trajectory cost $\mathcal{C}^{\text{sub}}$:
\begin{equation}
    \begin{aligned}
    \boldsymbol{\tau}^{\text{sub}} \sim \arg \min_{\boldsymbol{\tau} \in \mathbb{T}} \mathcal{C}(f_\theta, \boldsymbol{\tau})
    \end{aligned}
\end{equation}
where $\boldsymbol{\tau}^{\text{sub}}$ is an approximation to the optimal solution $\boldsymbol{\tau}^*$. 
With the trajectory $\boldsymbol{\tau}^{\text{sub}}$, the corresponding cost $\mathcal{C}^{\text{sub}}$ of the TO objective, together with the task-level loss $\mathcal{L}^{\text{sub}}$, are propagated back to optimize the network parameter $\theta$ using gradient decent and solves the BLO problem iteratively:
\begin{equation}
    \theta_{i+1} \leftarrow \theta_i - \omega \nabla_{\theta_i} \mathcal{F}(f_{\theta_i}, \boldsymbol{\tau}^{\text{sub}}_i),
\end{equation}
where $\omega$ is a hyperparameter of the learning rate. The gradient of the training loss $\mathcal{F}$ to the network parameter $\theta$ is denoted as follows:
\begin{equation}
    \nabla_{\theta}\mathcal{F} =  \left( \frac{\partial\mathcal{F}}{\partial f} \,+\, \frac{\partial \mathcal{C}}{\partial f} \right) \frac{\partial f}{\partial \theta} + \frac{\partial \mathcal{C}}{\partial \tau}\, \frac{\partial \tau}{\partial \theta}.
\end{equation}

\section{Experiments}

We assess our method through both simulated and real-world evaluations, comparing it to classic and learning-based baselines. The simulations are run on a 2.6GHz i9 laptop with an NVIDIA RTX 3080 GPU. Our real-world tests are carried out using the ANYmal legged robot~\cite{hutter2016anymal} equipped with an NVIDIA Jetson Orin for the execution of our planning method.

The classic motion primitives (\textbf{MP}) planner, as proposed by \citet{zhang2020falco}, integrates with a modularized pipeline~\cite{cao2022autonomous} and uses a 360$^\circ$ view LiDAR. It is considered the state-of-the-art (SOTA) non-learning method in terms of success rate and efficiency, serving as a performance reference for learning methods. As learning-based approaches, the supervised learning (\textbf{SL}) method by \citet{loquercio2021learning} and the reinforcement learning (\textbf{RL}) method by \citet{hoeller2021learning} serve as baselines. Both our method and the learning baselines use a front-facing stereo-depth camera with a frame rate of 15Hz. The RL method requires a downward-tilted camera view of 30$^\circ$, while the SL method requires a forward-looking camera. Our method is tested with both camera settings using the same policy weights.

Our method is trained using a combination of data collected in both simulated and real-world environments. We gather approximately 20k depth images from various camera positions in the Matterport3D~\cite{Matterport3D} environment as well as in simulated campus and tunnel~\cite{cao2022autonomous} environments using Gazebo~\cite{koenig2004design}. Additionally, we collect 10k images in real-world environments to adjust the policy for real-world perception noise. An example of depth observation in a simulated environment versus a real-world environment can be seen in Fig.~\ref{fig:image_examples}. The training takes about 20 hours (100 epochs) using 30k images and a single NVIDIA 3090 Ti GPU, without pre-trained ResNet weights. After just 20 epochs (4 hours), the policy can demonstrate a high success rate in navigating through the unseen simulated environment, as shown in Fig.~\ref{fig:training_plot}.

\begin{figure}
    \centering
    \includegraphics[width=0.9\linewidth]{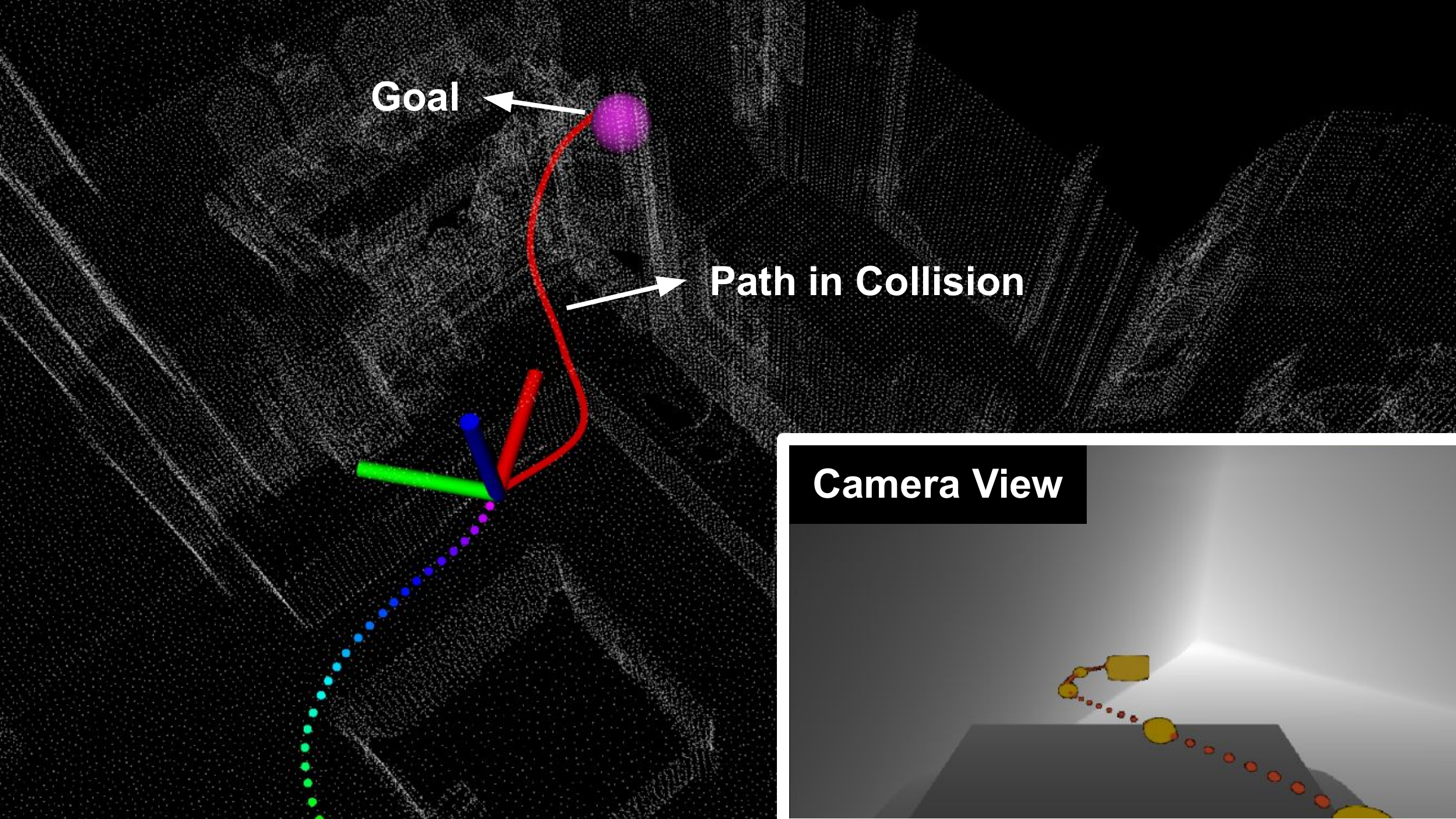}
    \caption{An illustration of a local minimum scenario. The point cloud, in white, depicts the surrounding environment.  The robot is trapped in a corner where no collision-free path to the goal (purple) exists within the FOV. The network predicts a path (red) with a high collision probability, which triggers the protective behavior to stop the robot.}
    \label{fig:local_minimal}
\end{figure}

\begin{figure}
    \centering
    \subfigure[]{\includegraphics[height=0.3\linewidth]{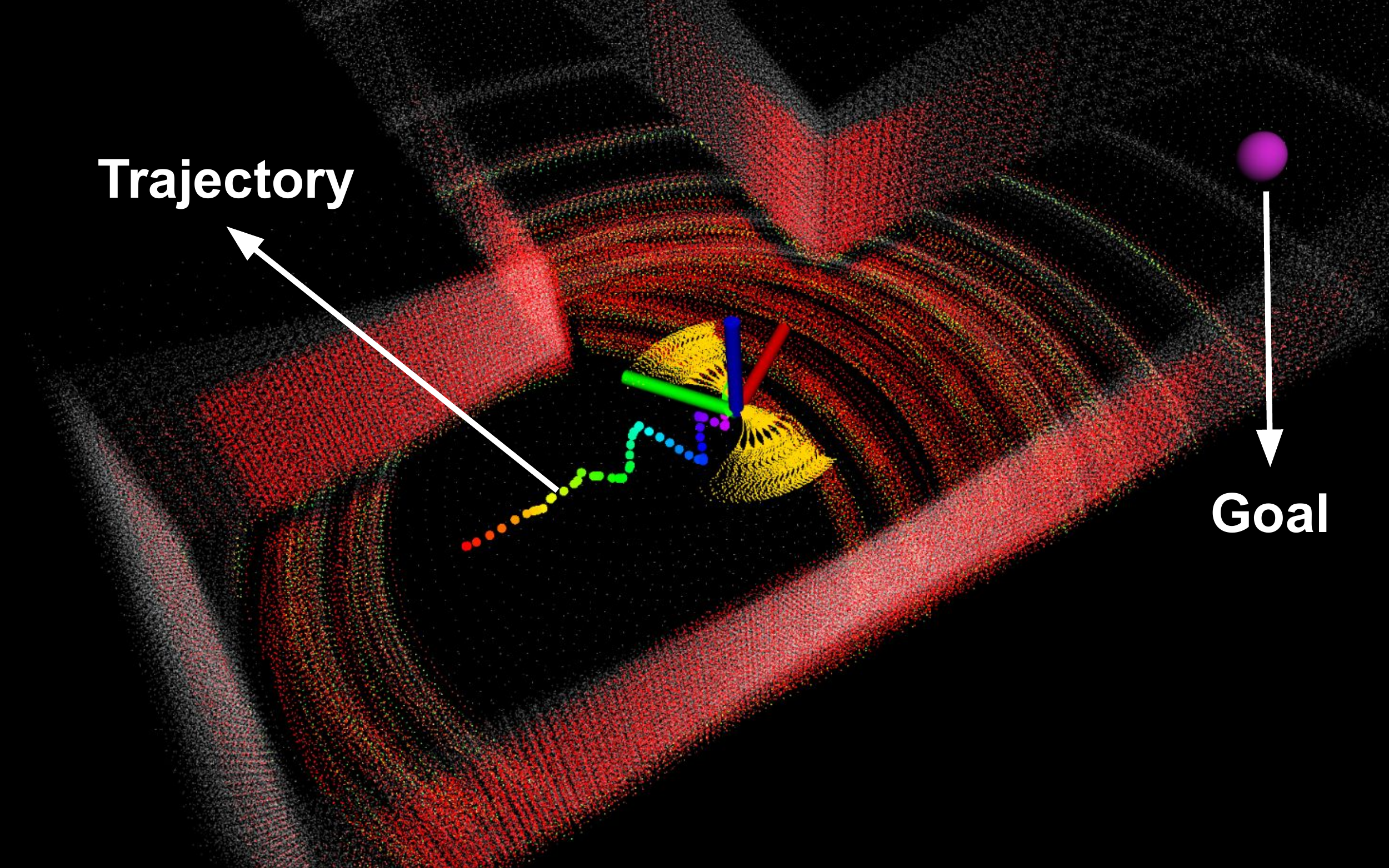}}
    \subfigure[]{\includegraphics[height=0.3\linewidth]{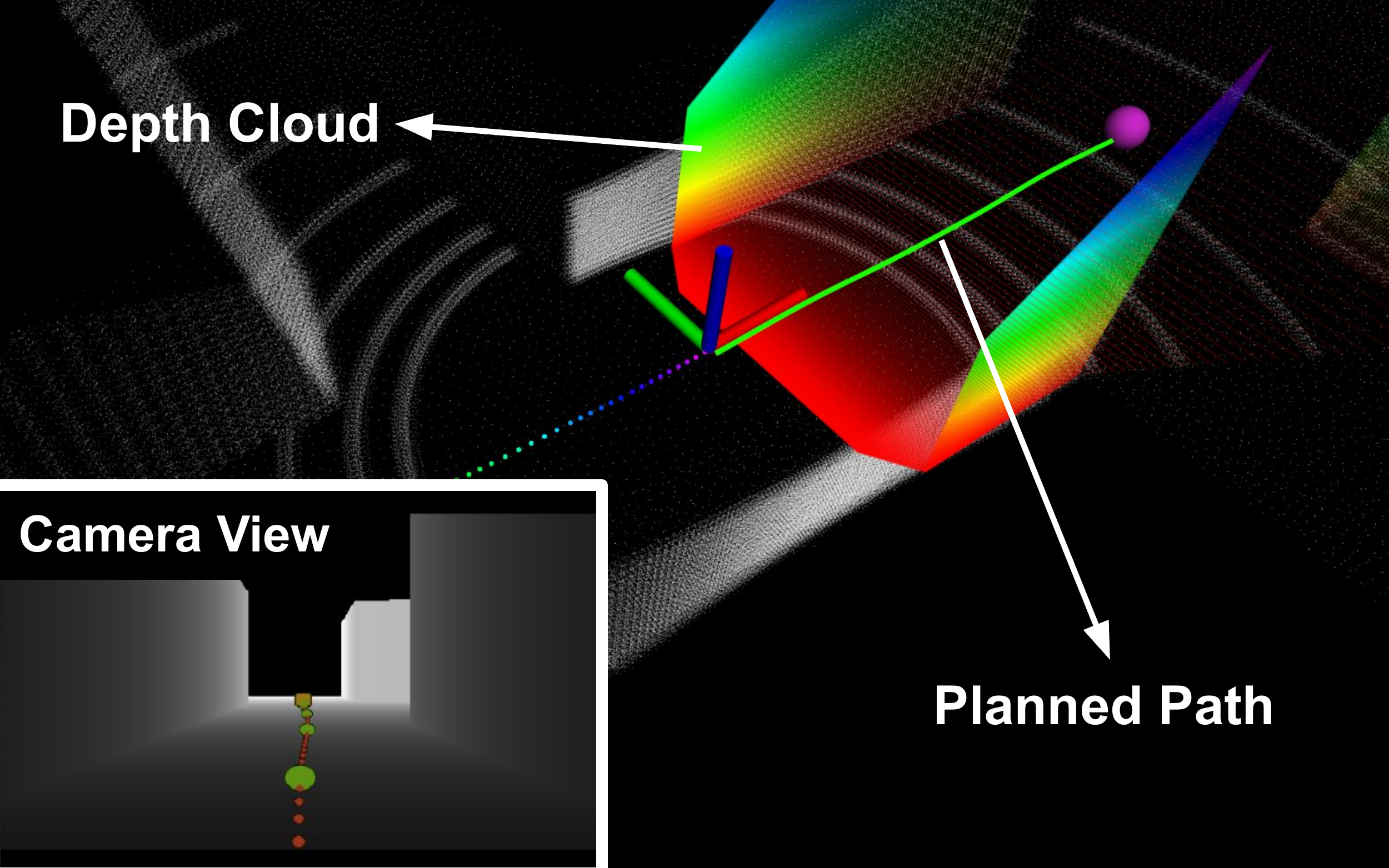}}
    \caption{Experiment against localization errors. (a) illustrates the compounding effect of the classic MP method with the modularized pipeline. The point cloud in red depicts the obstacles detected by the terrain analysis module. (b) shows the planning result of our method. In the same conditions, a feasible path (green) to the goal is planned based on the current depth observation displayed in the camera view image.}
    \label{fig:coumpouding_error}
\end{figure}

\begin{table}
\renewcommand{\arraystretch}{1.2} \centering{\scriptsize
\caption{Planning Performance across Four Types of Environments}
\label{tab:spl_table}
\begin{tabular}{cccccc}
\hlineB{3.5}
\multicolumn{6}{c}{SPL \% - Success weighted by Path Length} \\ \hline
        & Forest         & Garage         & Indoor         & Matterport     & Overall   \\ \hlineB{2.5}
\multicolumn{1}{l|}{MP~\cite{zhang2020falco} (LiDAR)}    & 95.09          & \textbf{89.42} & 85.82  & \multicolumn{1}{c|}{74.18}          & 86.13         \\
\hline
\multicolumn{1}{l|}{SL~\cite{loquercio2021learning}} & 65.58          & 46.70          & 50.03          & \multicolumn{1}{c|}{28.87}          & 47.80          \\
\multicolumn{1}{l|}{RL~\cite{hoeller2021learning}  (Tilt)}    & 95.08 & 69.43          & 61.10          & \multicolumn{1}{c|}{59.24}          & 71.21          \\

\multicolumn{1}{l|}{Ours  (Tilt)}  & 95.66          & 73.49          & 68.87 & \multicolumn{1}{c|}{76.67} & 78.67           \\
\multicolumn{1}{l|}{Ours}  & \textbf{96.37}          & 88.85          & \textbf{90.36} & \multicolumn{1}{c|}{\textbf{82.51}} & \textbf{89.52}           \\
\end{tabular}}
\end{table}

\begin{table}
\renewcommand{\arraystretch}{1.1} \centering{\scriptsize
\caption{Planning Latency in [ms] for Different Methods}
\label{tab:latency_table}
\begin{tabular}{ccccc}
\hlineB{3.5}
SL~\cite{loquercio2021learning}          & RL~\cite{hoeller2021learning}                  & MP~\cite{zhang2020falco}          & Ours      & Ours (\textit{Jetson}) \\ \hline
13.5 ($\pm$1.8) & \textbf{3.6 ($\pm$0.3)} & 24.9 ($\pm$4.2) & 5.9 ($\pm$0.4) & 11.4 ($\pm$0.6)    \\ \hlineB{2.5}
\end{tabular}}
\end{table}

\begin{figure*}
    \centering
    \includegraphics[width=0.99\linewidth]{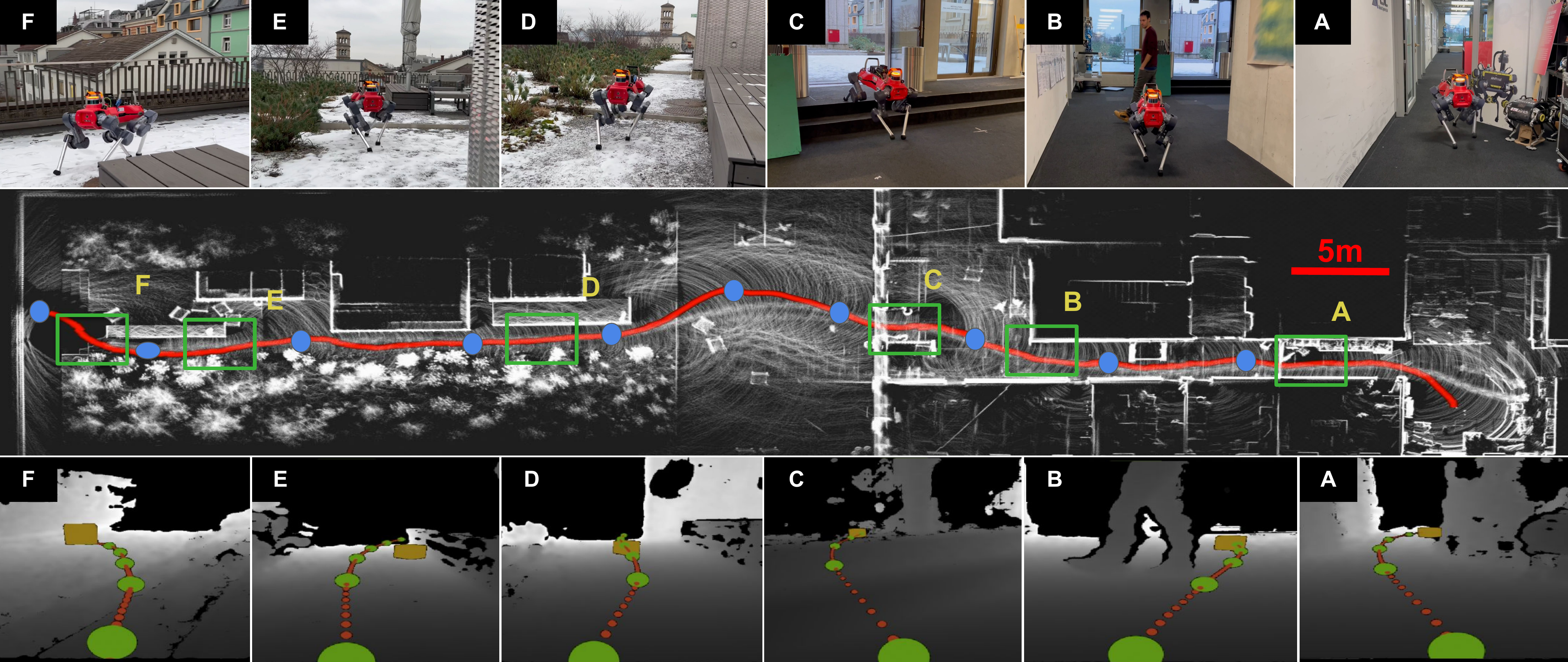}
    \caption{Real-world experiment with the legged robot, with the red curve indicating the robot's trajectory starting from right to left. The robot begins inside a building and navigates to the outside. Green boxes A-F mark different planning events during the route. The robot follows a series of waypoints (shown in blue) and plans in between to: (A) pass through doorways, (B, D, E) circumvent static/dynamic obstacles, and (B, F) ascend and descend stairs.}
    \label{fig:real_experinments}
\end{figure*}

\subsection{Simulation Experiments}

The simulated experiments are conducted in four different types of environments: indoor, garage, forest, and a Matterport3D room, as shown in Fig.~\ref{fig:planning_examples}, using the Autonomous Exploration Development Environment~\cite{cao2022autonomous}. The testing environments are not seen by the learning-based planning policies during training. In each environment, 30 pairs of start and goal positions are selected in traversable areas and given to the robot. The robot uses onboard sensing to search for collision-free paths and navigate to reach the goal.

Table.~\ref{tab:spl_table} presents the results of experiments. The performance of the planning methods is evaluated using SPL~\cite{anderson2018evaluation} (Success weighted by Path Length). In the experiments, the SL method is found to be inferior in generalizing to the four unseen testing environments. On the other hand, the RL method performs well in the forest but struggles in the other three environments. This is due to its overfitted perception header trained with a downward-tilted camera, resulting in limited field-of-view (FOV) for planning effectively in complex environments. Furthermore, the perception header of the RL policy is trained using a self-supervised approach, not specifically optimized for planning. In contrast, our method demonstrates consistent and high performance across all different types of unseen environments and outperforms even the classic method with a 360$^\circ$ degree FOV LiDAR. Furthermore, our policy is capable of directly generalizing to different camera settings. Even with the same downward-tilted camera as the RL method, our policy can still achieve the best performance among all learning-based methods, shown in Table~\ref{tab:spl_table}. On the other hand, the robot's performance can be impacted by local minimal scenarios when its FOV is limited. For those cases, our network predicts ``fear" values to detect trajectory collisions and stop the robot before crashing, as demonstrated in Fig.~\ref{fig:local_minimal}. In summary, our proposed planner has the capability to achieve, on average, 87$\%$ better performance than the SL method and 26$\%$ better performance than the RL method in terms of SPL to reach destinations.

The MP approach models the environment using a metric-based method and offers superior performance and generalization compared to learning-based baselines~\cite{loquercio2021learning, hoeller2021learning} as shown in Table~\ref{tab:spl_table}. However, its modularized pipeline can result in high latency and sensitivity to noise, e.g., localization errors. To assess this, we add random localization noise with a standard deviation of 5cm to the system. This causes false-positive obstacle detections on the ground due to compounding effects between the mapping and terrain analysis modules, as shown in Fig~\ref{fig:coumpouding_error}. With these false detections, the MP planner cannot find a ``collision-free" path to the goal. Our proposed method demonstrates high robustness, even in the presence of random localization noise, through its end-to-end pipeline that takes instantaneous sensor input for planning. Despite not using a panoramic sensor like LiDAR, our method offers similar generalization and even better performance than the classic MP method. Table~\ref{tab:latency_table} shows our method's computational efficiency and small planning latency, which is more than 4 times faster than the MP method but slightly slower than the RL approach.


\subsection{Real-World Experiment with Legged Robot}
Our experiment evaluates the effectiveness of our method in real-world scenarios using the legged robot ANYmal~\cite{hutter2016anymal}. The robot is equipped with an Intel Realsense D435 front depth camera for depth measurement. The experimental setup involves both dynamic and static obstacles, with the robot navigating from indoor to outdoor environments following sequential waypoint commands from a human operator, as shown in Fig.~\ref{fig:real_experinments}. The planner navigates the robot through doorways, around both dynamic and static obstacles, and up and down stairs. The indoor and outdoor environments have different lighting conditions, which results in varying levels of noise in the depth measurements. Here, our planner relies on a localization method~\cite{khattak2020complementary} to translate the goal into the robot frame for the planning network. 

\begin{figure}
    \centering
    \subfigure[Underground]{\includegraphics[height=0.3\linewidth]{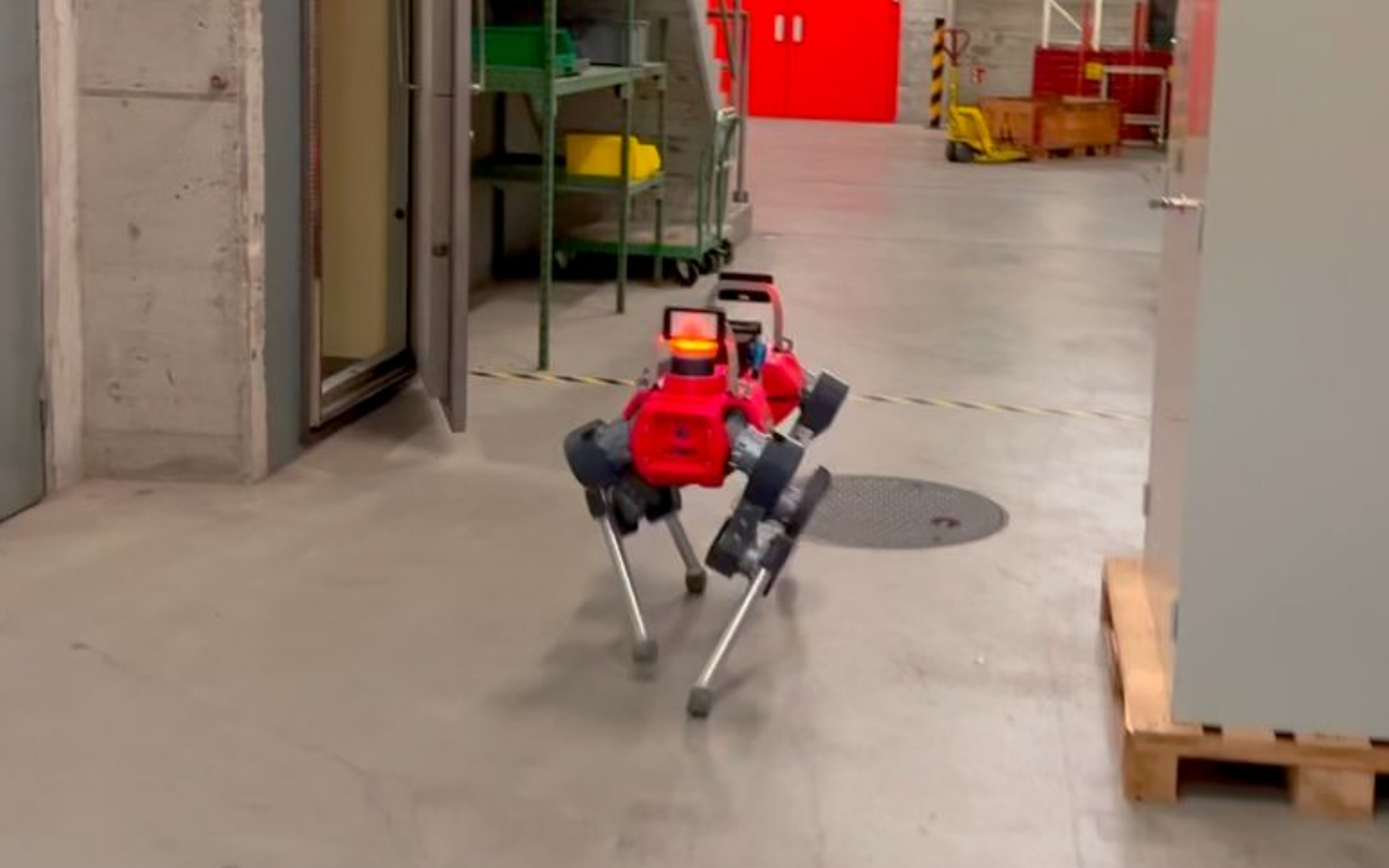}}
    \subfigure[Laboratory]{\includegraphics[height=0.3\linewidth]{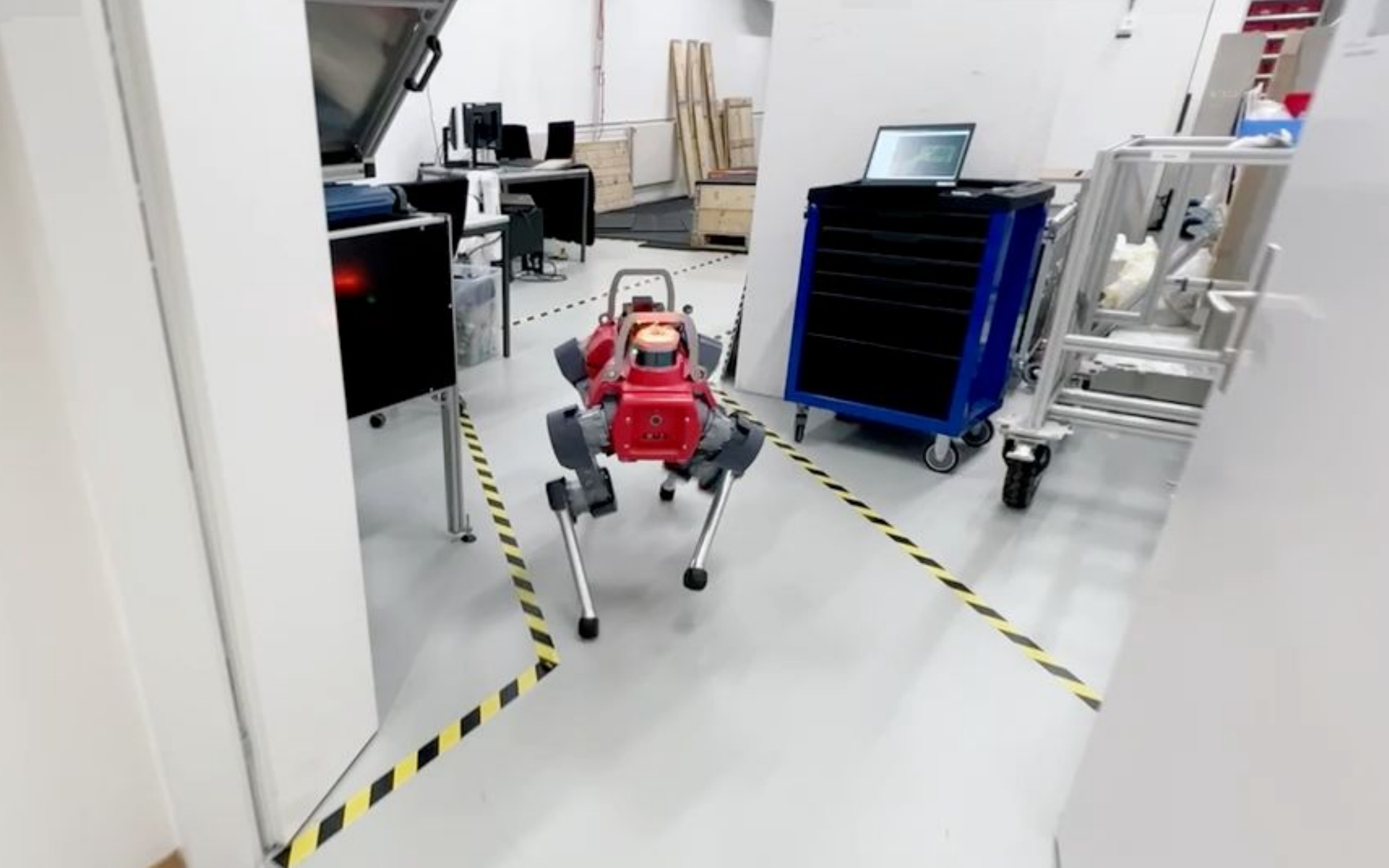}}
    \subfigure[Human-made ``Maze"]{\includegraphics[height=0.3\linewidth]{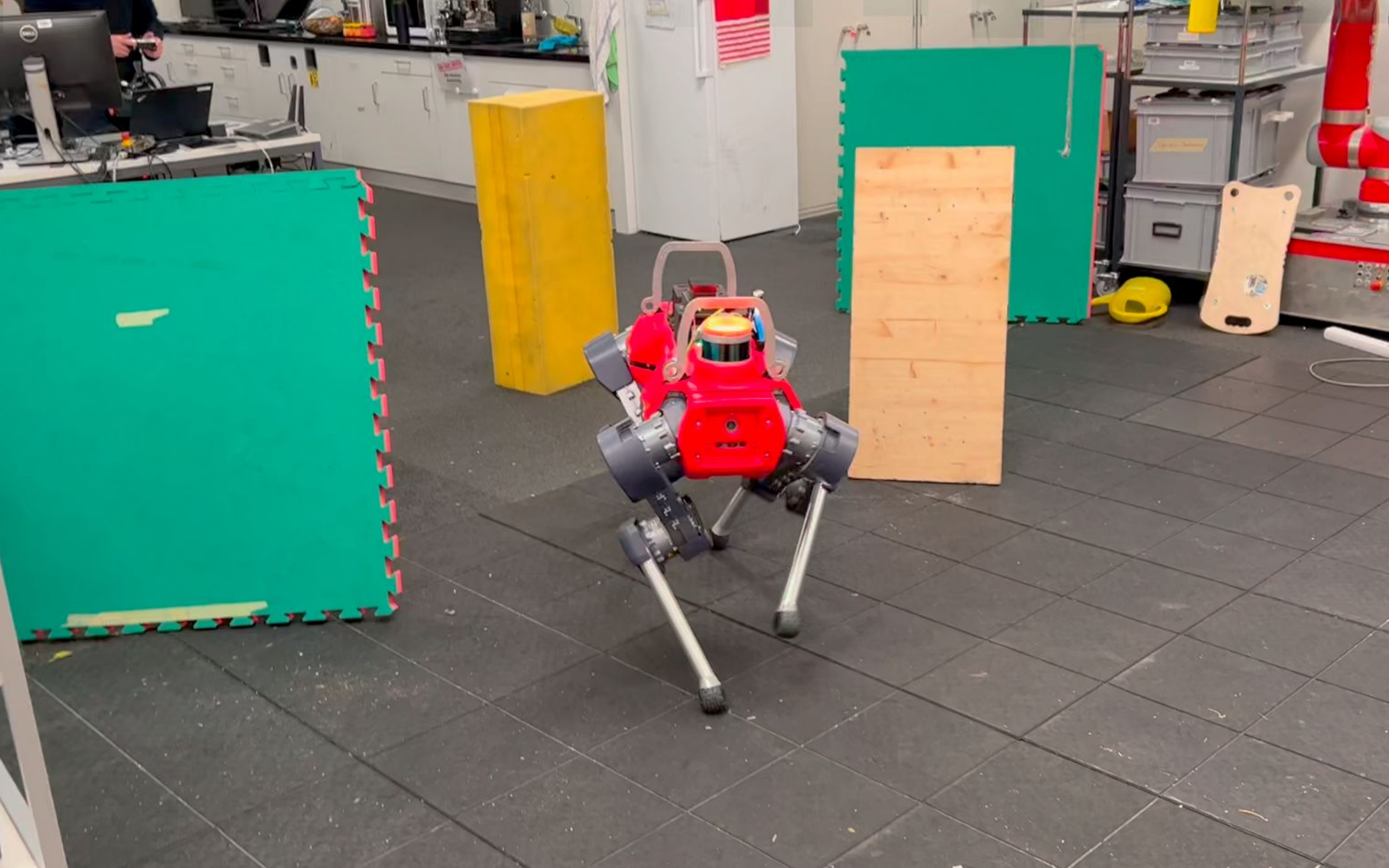}}
    \subfigure[Terrace]{\includegraphics[height=0.3\linewidth]{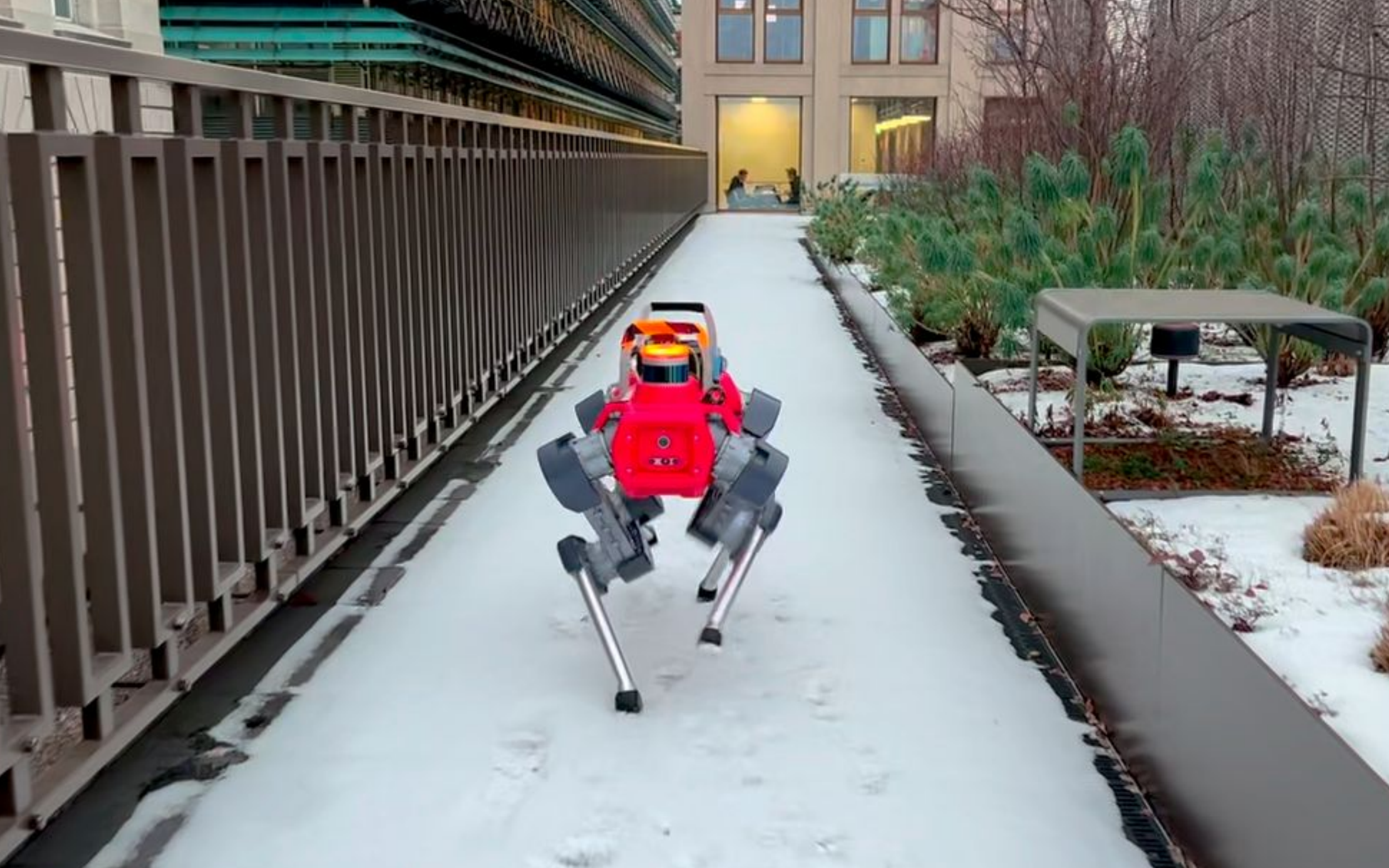}}
    \caption{Planning method testing in different unseen environments, indoor and outdoor, with various obstacle setups and lighting conditions.}
    \label{fig:real_experinments_2}
\end{figure}

We also conducted experiments in various environmental settings to assess the generalization and efficiency of our planning policy, as illustrated in Fig.~\ref{fig:real_experinments_2}. In human-made mazes, the goal may be hidden from the start, requiring the robot to navigate around obstacles. Outdoors, the robot must traverse various terrains, such as snow, grass, and plants, to reach its destinations. In the underground environment, the robot encounters different lighting conditions and unfamiliar obstacle shapes. Despite these challenges, our method guides the robot successfully to its destinations while maintaining a low average latency of 11.4ms on the Nvidia Jetson Orin onboard computer. Furthermore, our method, which relies solely on a single depth frame, can operate independently of the localization module if the goal is set in the robot frame. We use a joystick command to set the goal in the robot frame as directional guidance, and the method then calculates safe paths in the local frame for the robot to follow.

\section{Conclusion}

In this paper, we present an end-to-end planning framework based on a novel imperative learning (IL) approach. The method involves a bi-level optimization (BLO) process that combines network update and metric-based trajectory optimization during training to produce smooth and collision-free trajectories using only a single depth measurement. The IL is able to utilize task-level loss and optimize through direct gradient descent. This allows the method to be trained in an efficient unsupervised manner, eliminating the need for explicit trajectory labels. In the experiment, we benchmark the performance of our planner with the SOTA classic non-learning method~\cite{zhang2020falco} and learning baselines~\cite{loquercio2021learning, hoeller2021learning}. Our experiments demonstrate that the resulting policy has the capability to achieve efficient planning and generalize to various unseen environments compared to previous approaches. Further, we evaluate our method in various real-world settings, including dynamic environments and different camera settings. The results indicate the effectiveness of our method in real-world deployment, generalizing to novel environments and camera configurations, and being robust against perception and localization errors.


\section{Discussion}
This paper presents a novel framework for training a perceptive planning policy that relies solely on depth measurement as input. While RGB images can offer supplementary information beneficial for planning, images collected from simulation may result in a significant difference between real environments. Thus, further research is required. Additionally, we realize that adding memory structure to the perception network could have improved the performance of the planning in the static environment. However, utilizing the current pipeline, the memory structure may compromise the safety of the planning to avoid dynamic obstacles. To address this issue, it may be necessary to incorporate a time-variable cost map in formulating the training loss. Overall, this work aims to demonstrate the innovative concept of using the imperative learning approach, and the result can be considered preliminary.

\section{Acknowledgement}
This research was supported by: the Swiss National Science Foundation (SNSF) under project 188596, the National Centre of Competence in Research Robotics (NCCR Robotics), and the European Union's Horizon 2020 research and innovation program via grant agreements No. 101016970, No. 101070405, and No. 852044. Additionally, this work also benefited from the ETH Zurich Research Grant.

{
\small
\bibliographystyle{plainnat}
\bibliography{references}
}

\end{document}